\documentclass{article}
\usepackage[margin=2.2cm]{geometry}
\usepackage[utf8]{inputenc}
\usepackage{graphicx}
\usepackage{color}
\usepackage{subcaption}
\usepackage[hidelinks]{hyperref}
\usepackage{multirow}
\usepackage[font=sf]{caption}
\usepackage{amsmath} % assumes amsmath package installed
\usepackage{amssymb}  % assumes amsmath package installed
\usepackage[font={footnotesize}]{caption}
 \usepackage{siunitx}

\graphicspath{ {./images/} }
\usepackage[
backend=biber,
style=numeric,
sorting=none
]{biblatex}
\newcommand*{\figref}[2][]{%
  \hyperref[{fig:#2}]{%
    Figure~\ref*{fig:#2}%
    \ifx\\#1\\%
    \else
      \,#1%
    \fi
  }%
}

\newcommand*{\tableref}[2][]{%
  \hyperref[{table:#2}]{%
    Table~\ref*{table:#2}%
    \ifx\\#1\\%
    \else
      \,#1%
    \fi
  }%
}

\newcommand*{\equref}[2][]{%
  \hyperref[{eq:#2}]{%
    Equation~\ref*{eq:#2}%
    \ifx\\#1\\%
    \else
      \,#1%
    \fi
  }%
}

\begin{document}
\section*{\textsf{Markerless 3D human pose tracking through multiple cameras and \\AI: Enabling high accuracy, robustness, and real-time performance}}

% \section*{\textsf{Markerless human pose tracking in the wild: A 3D real-time \\approach based on multi-camera AI-driven human pose estimation}}
% \section*{Real-time, markerless 3D human pose tracking in the wild: A multi-camera, AI-driven human pose estimation approach}
% \section*{Real-Time MarkerLess 3D Human Pose Tracking\\ Exploiting Multi-Camera Views and AI-Driven Pose Estimation}
\subsubsection*{\textsf{Luca Fortini$^{*1,2}$, Mattia Leonori$^1$, Juan M. Gandarias$^1$, Elena de Momi$^2$,  Arash Ajoudani$^1$}}
$^1$Human-Robot Interfaces and Interaction, Istituto Italiano di Tecnologia, Genoa, Italy.\\
$^2$Department of Electronics, Information and Bioengineering, Politecnico
di Milano, Milan, Italy.\\
$^*$Corresponding author: \href{mailto:luca.fortini@iit.it}{luca.fortini@iit.it}

% SECTION 1: Abstract
\vspace{0.3cm}
\label{Sec::abstact}
\noindent
\textsf{\textbf{Tracking 3D human motion in real-time is crucial for numerous applications across many fields. Traditional approaches involve attaching artificial fiducial objects or sensors to the body, limiting their usability and comfort-of-use and consequently narrowing their application fields. Recent advances in Artificial Intelligence (AI) have allowed for markerless solutions. However, most of these methods operate in 2D, while those providing 3D solutions compromise accuracy and real-time performance.
To address this challenge and unlock the potential of visual pose estimation methods in real-world scenarios, we propose a markerless framework that combines multi-camera views and 2D AI-based pose estimation methods to track 3D human motion. Our approach integrates a Weighted Least Square (WLS) algorithm that computes 3D human motion from multiple 2D pose estimations provided by an AI-driven method. The method is integrated within the Open-VICO framework allowing simulation and real-world execution. Several experiments have been conducted, which have shown high accuracy and real-time performance, demonstrating the high level of readiness for real-world applications and the potential to revolutionize human motion capture.}}

\vspace{0.8cm}

% SECTION 2: Introduction
% \section{Introduction}

\label{Sec::introduction}
% GENEARL BACKGROUND AND MOTIVATION
In recent years, human motion capture has emerged as an essential instrument for multiple and diverse applications such as sports~\cite{fastovets2013athlete}, robotics~\cite{ott2008motion}, animation~\cite{pan2011sketch}, and rehabilitation~\cite{zhou2008human}, to name just a few. Accurately capturing and replicating human motion enables more realistic and natural movements in virtual environments~\cite{chan2010virtual}, allowing its integration within applications in the metaverse~\cite{yang2022metafi} and digital twins~\cite{saddik2018digital}. A recent example of the latter is the last release of the Nvidia Omniverse~\footnote{\url{https://www.nvidia.com/en-us/omniverse/}}, which includes animated human characters. There is no doubt that human motion capture has enormous potential to improve people's lives, whether for performance enhancement and safety of workers or athletes, or integration into everyday applications for safety and entertainment.

% TRADITIONAL EXISTING SOLUTIONS
At present, human motion capture or human pose tracking is mainly carried out using marker-based systems. Such solutions use physical markers attached to the body that are tracked by cameras or other sensors~\cite{vanderkruk2018accuracy}. Their primary attribute is their high accuracy, especially the optical-based maker-based systems, for which there are commercial solutions such as Vicon~\footnote{\url{https://www.vicon.com/}} or Optitrack~\footnote{\url{https://optitrack.com/}}. IMU-based systems that integrate multiple Inertial Measurement Unit sensors in different parts of the body are the other flagship example of marker-based (motion capture) mocap systems, of which there are also commercial solutions such as Xsens~\footnote{\url{https://www.movella.com/}}. Although the latter does not have the same performance in terms of precision as optical-based systems, it is widely used in many applications where it is even considered as ground truth~\cite{gandolla2020wearable}. Although the marker-based approach is the most used, it has some widely known drawbacks and limitations~\cite{mundermann2006evolution}. The guaranteed high accuracy comes with non-negligible hinder of the user, intra-session variability due to the need to place the markers in the exact body places, or battery life for active sensors compromising their application in many scenarios~\cite{menolotto2020motion}. Moreover, the optical systems' measurements might be affected by line-of-sight effects. Correspondingly, the position accuracy of IMU-based systems is affected by drift over time as they require double integration of linear acceleration, needing frequent calibration.

% CHALLENGES AND PROBLEM STATEMENT
Apart from marker-based solutions, an accurate and real-time markerless human pose-tracking system that overcomes the aforementioned issues would revolutionize human motion capture. Allowing humans to only use their natural movements in real-time without attached markers to control virtual or augmented reality environments, to interact physically with real machines, or to monitor human motion would unlock countless possibilities for a wide variety of fields. These include gaming, entertainment, healthcare, sports, ergonomics, and other domains where online streaming, quantification and analysis of natural human movements are critical, providing a more intuitive and natural means of communication between humans and machines that could enable more seamless and efficient interactions in a wide range of settings.
Recent advances in markerless monocular skeleton detection have enabled new applications requiring semi-accurate body parts tracking~\cite{colyer2018review}. Typically, they involve the inference of joint coordinates and the reconstruction of a human skeletal representation. In the last few years, 2D human pose estimation reached astounding detection rates on all different human joints~\cite{munea2020progress}. The most widespread methods are OpenPose~\cite{cao2021openpose}, MediaPipe~\cite{lugaresi2019mediapipe}, or AlphaPose~\cite{fang2022alphapose}. This progress has been made possible in large part by the success of Convolutional Neural Networks (CNNs) and the emergence of large-scale accessible datasets. However, only recently these new architectures have been deployed to tackle similar problems in 3D. The main challenge for these new 3D markerless pose estimation methods is to be competitive against classical marker-based motion capture systems. The ultimate goal would be a complete and accurate 3D reconstruction in real-time of an individual's motion from simple monocular images with tolerance to severe occlusions~\cite{chen2020monocular}. 

% RELATED WORKS (GENERAL)
There are multiples way to tackle the problem of markerless 3D human pose estimation. Some approaches considered traditional computer vision techniques, i.e., not using learning or AI-based methods~\cite{puwein2015joint, ballan2008marker}. Some others consisted of pure learning-based methods~\cite{park2016tracking, kidzinski2020deep, luvizon2022consensus}. Apart from the methodological approach, research studies have considered reconstructing from monocular images~\cite{agarwal2005monocular}, while others employ Stereo Cameras~\cite{azad2007stereo} or RGB-D sensors~\cite{kohli2013key}. Despite the acceptable results obtained by these works, none has achieved outcomes comparable to marker-based solutions. In addition, many require complicated calibration processes, training, or only work in highly structured environments, such as laboratory settings with monochromatic backgrounds. Another common drawback with many of these works is that they require post-processing work or present high latency, preventing their use in real-time applications. A thorough and detailed state-of-the-art review of markerless tracking systems can be found in~\cite{desmarais2021review}.

% SIMILAR WORKS (PARTICULAR APPROACH)
One promising approach involves exploiting 2D AI-based skeleton trackers (e.g., OpenPose) to reconstruct a 3D representation of the human body~\cite{garcia2019human}, benefitting the excellent outcomes of recent AI methods that have proven in extracting features and classification from images. A common and straightforward approach uses 3D sensors as RGB-D cameras~\cite{lamon2020visuo, liu2022simple}. However, as they reconstruct the 3D data from methods that usually work at the pixel level, even if they present excellent results at this level, the accuracy drastically drops when extrapolated to the 3D world.

These issues can be addressed with a methodology explored previously in the literature that uses 2D AI-based human skeleton tracking methods to obtain the key points (i.e., joint positions) and perform a triangulation to obtain a 3D skeleton that is robust and accurate. A method using two cameras and OpenPose for gait analysis was presented in~\cite{zago2000tracking}. Although valid for the proposed application, this method shows errors up to 6cm without considering occlusions or other difficulties that could be large for other applications. In addition, the cameras are arranged in a plane parallel to the frontal plane of the person, which facilitates calibration and promotes accuracy at the cost of sacrificing its potential applicability in many applications. A similar solution is presented in~\cite{nakano2020evaluation}. In this work, five cameras and OpenPose were used with a triangulation method based on Direct Linear Transformation. However, this is a human motion capture rather than human pose tracking since all videos are recorded first and post-processed afterward, which makes it impossible to use in real-time applications. Also, details should be included on the calibration method for obtaining the extrinsic parameters of the cameras, and more details should be given on the implementation, which makes it difficult to replicate the work. In addition to these works, a more complex triangulation algorithm with OpenPose and multiple cameras is presented in~\cite{slembrouck2020multiview}. Although the authors claim that the method can be used in real-time using four cameras with 2 Nvidia RTX 1080 GPUs, no evidence shows that the system works in real-time. Some results are shown in terms of frame identifier and not time, which suggests that the data has been subsequently analyzed frame by frame and not in real-time. Finally, the work presented in~\cite{izaak2022accuracy} conducted an experimental evaluation of three different modalities of 3D markerless estimation. However, as in previous works, no evidence is shown that the system works in real-time, no details are given on the calibration of the multi-camera system nor the implementation of the methodology, and the results are presented at the joint level. All this makes it hard to replicate and impossible to compare the results with other works. These works proposed partial solutions to some drawbacks of marker-based systems but presented some limitations revealing challenges yet to be overcome. 

\begin{figure}[!t]
    \centering
    \includegraphics[width=0.85\columnwidth]{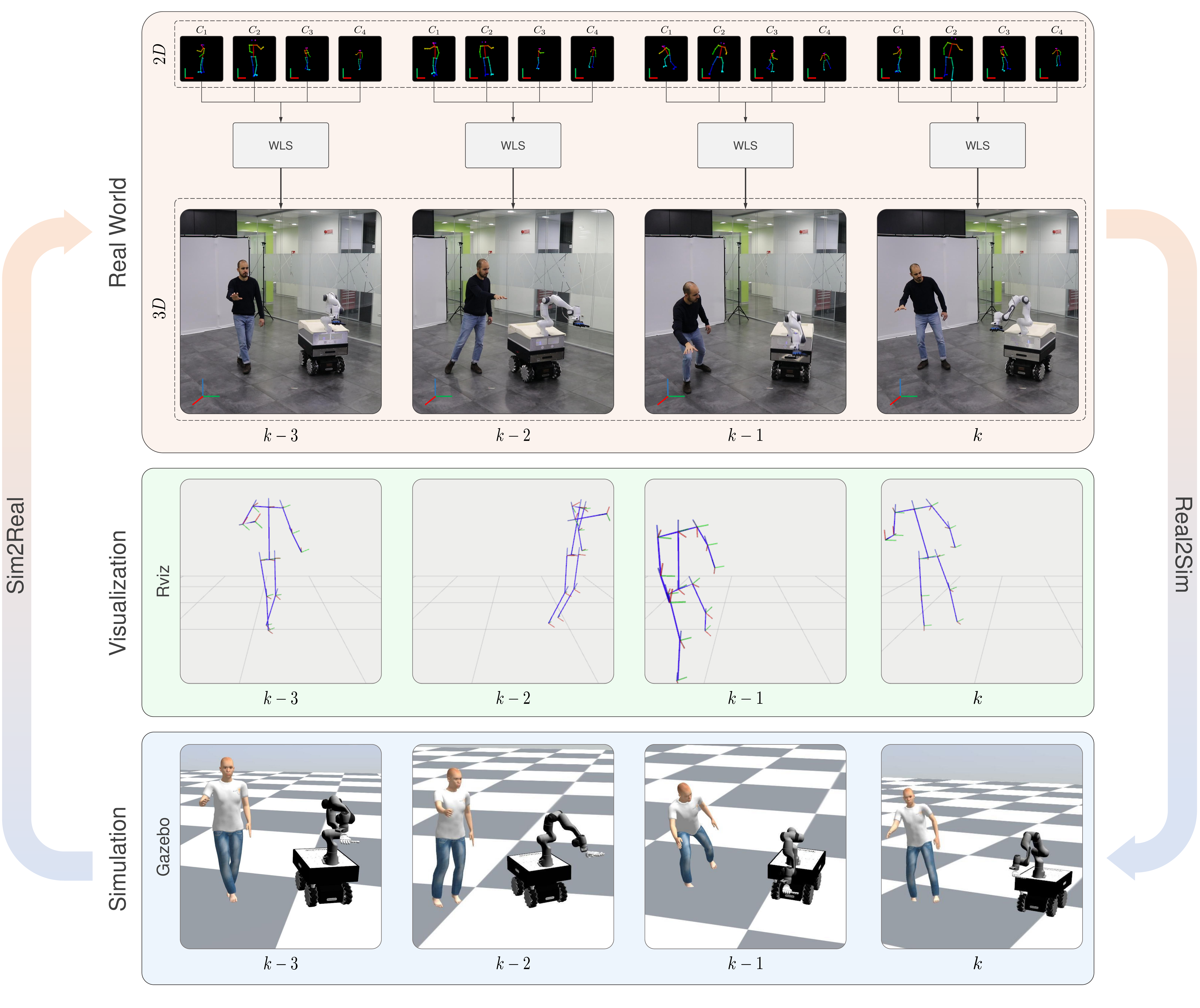}
    \captionsetup{labelfont=bf,font=sf}
    \caption{\textbf{A practical application to show the framework's pipeline.} In the picture are shown four shots of a teleoperation task using the 3d reconstruction of the hand to move the end-effector of a mobile manipulator (MOCA). From top to bottom we can see the proposed framework offers the possibility to loop between the real world and the simulation thanks to Open-VICO. The 2D skeleton single views of the cameras are fused to get a 3D representation of the human to maneuver the end-effector. The blue skeletons are the result of the reconstruction and visualization feedback in RViz. The trajectories can be exported and repurposed to animate a human-like model for simulation analysis in Gazebo.}
    \label{fig:digest}
\end{figure}

% CONTRIBUTION
Inspired by the potential of this approach and by the challenges that remain to be solved, this work contributes to the problem of real-time and 3D human pose tracking with a comprehensive, unified framework that integrates the following characteristics meeting many of the challenges yet to be solved in this field. \figref{digest} illustrates a representative of our framework. 

\begin{itemize}
	\item First, we propose a markerless approach with a weighted, multi-score reconstruction method that uses an AI-based 2D human pose tracker to rebuild the skeleton of a person in 3D from multiple camera views using a Weighted Least Square (WLS) triangulation method. The method considers different factors that can affect the performance of the 3D reconstruction and ponderates them based on heuristics. 
 
    \item This method has been integrated within our ROS~\footnote{The \href{https://www.ros.org/}{Robot Operating System (ROS)} is a set of open-source software libraries and tools for building robotic applications.} and Gazebo~\footnote{\href{https://gazebosim.org/home }{Gazebo} is an open-source software tool inside the ROS ecosystem for simulating 3D environments that enables testing with robot designs, control algorithms, and sensors in virtual settings.}-based open source toolkit Open-VICO~\cite{fortini2022openvico}, forming a unified framework for real-time 3D human pose tracking. This integration renders the framework flexible in different meanings: i) it allows the integration of multiple AI-based 2D skeleton trackers, ii) it allows abstraction from the hardware point of view both in simulation and in the real world, thanks to the message-delivering management handled by ROS, iii) as illustrated in \figref{digest} by the \textit{Sim2Real} and \textit{Real2Sim} arrows, the framework allows a straightforward transfer from simulation to the real world environments and vice versa, iv) the use of ROS visualization tools as Rviz~\footnote{\href{Rviz}{http://wiki.ros.org/rviz} is a 3D visualization tool for ROS that allows users to visualize and interact with sensor data, robot models, and other objects.} within the framework allows monitoring multiple parameters of the tracking and the application both in the real world and in simulation.

    \item It is necessary to mention that the proposed framework aims to be used for human pose tracking, which differs from human motion capture in that the former must be done in real-time. While existing works propose partial triangulation or reconstruction solutions that can work in real time, the entire pipeline must work in real-time, considering the computational requirements of the 2D AI-based estimations. Thanks to a multi-GPU setup and the integration in ROS, our framework guarantees the real-time operation of the entire pipeline, as demonstrated in the next section and illustrated in \figref{digest} where the teleoperation of a robot is performed in real-time.

    \item A solution that follows this methodology to be used in the wild must also have a suitable calibration system that allows the number of cameras required for the application to be available in the environment and to carry out a single calibration in a simple way to obtain the extrinsic parameters of the multi-camera system, which must also be accurate. Our framework incorporates a simple and quick calibration routine based on the method presented in~\cite{sarmadi2019simultaneous}.

    \item As with any emerging technology, validation against an established gold standard is crucial to help researchers understand a novel system's strengths and weaknesses. Related works compared their results with benchmark marker-based motion capture solutions commercially available. However, there are some limitations related to marker positioning and the validity of experimental results between different sessions or even in the same session in which markers can move as a cause of the subject's movement. The sub-millimeter error claimed by the marker-based markers is no longer such if the markers are not always placed in the exact location. In addition, marker-based systems present other problems that preclude their use as a validation method in many applications (e.g., in case of occlusions). To the best of our knowledge, our framework, unlike the rest of the existing works, is the first and only one that allows comparing the results with real ground truth, thanks to its integration in a simulation environment, where the actual positions of the joints are controlled and known without any error.

\end{itemize}

% SECTION 3: Results
\section*{\textsf{Results}}
\label{Sec::results}

The proposed 3D reconstruction framework has been evaluated through qualitative and quantitative testing. The outcomes of these experiments are presented in the following subsections to analyze the framework's performance. More information about the methodology, materials and implementation is described in the \textit{Methods} section.

This section presents the overall outcomes of the framework and its features for real-world and simulation applications with real-time requirements. An example was already depicted in \figref{digest}. This figure shows how straightforward it is to include the proposed methodology in ROS within a real robotic application to teleoperate a wheeled mobile manipulator in real-time under the controller schema presented in~\cite{wu2021unified}. In the application, the mobile manipulator reproduces the motion of a subject. The teleoperation starts with acquiring the initial position of the user's right hand using the proposed methodology and the end-effector frame w.r.t. the shared world frame. The subsequent acquisitions are needed to compute the relative hand displacements from the initial position and send them to the robot controller as the desired end-effector position. Note that this application can only be conducted under a real-time control approach. 

The whole setup has also been tested in a simulation context.
Similarly, human movement can be recorded and repurposed in a Gazebo environment to virtually animate a human model. These operations fall within the concepts of \textit{Sim2Real} and \textit{Real2Sim}. Overall, both are essential concepts in robotics and AI research, as they help bridge the gap between simulated and real-world environments and enable the creation of more robust and accurate systems.

\begin{figure}[t!]
    \centering
    \includegraphics[width=0.85\columnwidth]{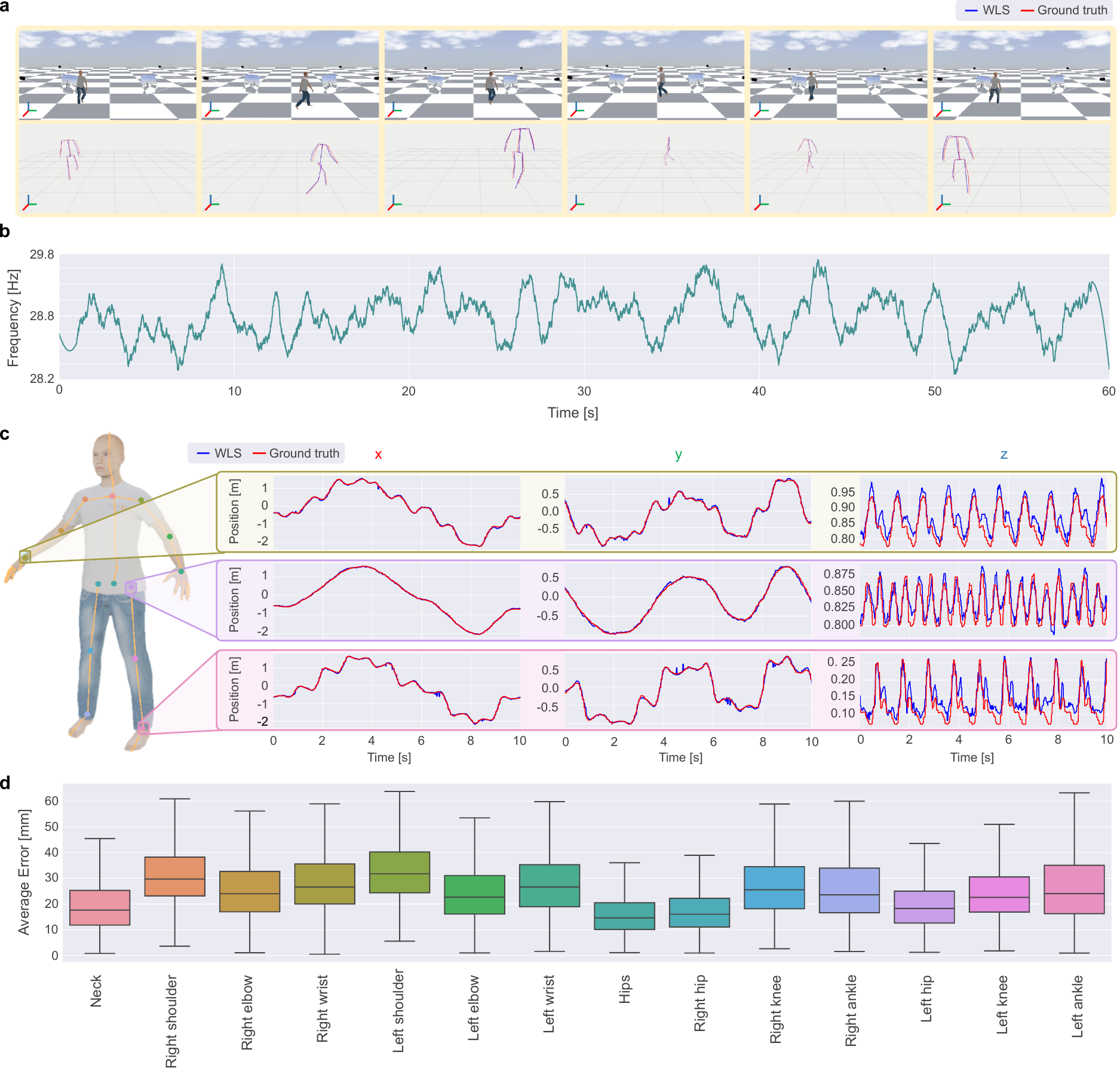}
    \captionsetup{labelfont=bf,font=sf}
    \caption{\textbf{Dynamic evaluation over a walking task in simulation}. Across all images from top to bottom the result of the proposed triangulation is indicated in blue while the groundtruth skeleton configuration in red. (\textbf{a}) Six screenshots randomly captured during the task. (\textbf{b}) Frequency variation of the triangulation algorithm over the task. (\textbf{c}) Sample trajectories of three explanatory landmarks (right wrist, left hip and left ankle) obtained from the reference groundtruth and the weighted triangulation. (\textbf{d}) Average positional error and dispersion between the groundtruth joints' positions and the estimated ones in mm.}
    \label{fig:simulation}
\end{figure}

\subsection*{\textsf{3D human pose tracking: Accuracy analysis}}

This section reports results on the performance of the proposed method in reconstructing 3D skeletons from RGB video sequences in simulation. 
In particular, we present the results of two experiments. The first is a dynamic experiment in a simulated environment without occlusions to measure the overall accuracy of the proposed method in a dynamic application (i.e., walking). The second experiment consists of a static environment. In this regard, we present two situations, one with occlusions and the other without them. This way, we expose the results regarding the robustness of the proposed method.

In \figref{simulation}a, a human-like character is simulated to walk randomly within a virtual workspace, which is captured by four cameras positioned at equally spaced points on a 4.5~\si{m} radius circumference. The simulated walking speed is set to 1.35~\si{m/s}, which can be considered an average speed of regular walking, and lasts 1 minute. A Gazebo plugin available in the Open-VICO toolbox makes it handy to retrieve the position of the underneath skeleton as ground truth. The bottom sequence shows six screenshots illustrating the overlapping of the resulting 3D triangulated skeleton using the WLS method (in blue) with the ground truth (in red). The plot in \figref{simulation}b displays the time performance of the method, dropping very few Frame Per Second (FPS) in comparison to the collecting frequency of the cameras, which was set at 30~\si{FPS}. 
The tracked position of three explanatory joints is depicted in \figref{simulation}c. The reader can appreciate that the proposed system accurately follows the ground truth positions. We use the Mean Per Joint Position Error (MPJPE), considered the most used metric to analyze 3D human pose estimation performance, to quantitatively evaluate the results averaged over all the sequences and the spread found in these measurements. MPJPE measures the mean Euclidean distance (i.e., the average L2 norm) of the difference between each joint's estimated position and ground truth in millimeters according to the following equation.  
\begin{equation}
\text{MPJPE}_j=\frac{1}{N_{f}} \sum_{n=1}^{N_f}\left\|p_{n,j}-\hat{p}_{n,j}\right\|_2,
\end{equation}
where $f$ denotes the frame index and $j$ denotes the corresponding joint. $\hat{p}_{n,j}$ is the estimated position of joint $j$, and $p_{n,j}$ is the corresponding ground truth. \figref{simulation}d reveals that the average errors per joint are all under 35~\si{mm}. The average error among all the joints is 25.83~\si{mm}, being the maximum the left shoulder with an average error of 33.07~\si{mm}, and the minimum the hips with an average error of 15.75~\si{mm}. These results also show small standard deviations, demonstrating the reliability and accuracy of the proposed framework for dynamic movements.

The second analysis evaluates the performance of the system considering the influence of the weights used to ponderate the WLS method (see section \textit{Methods}) on the final estimation. The experiments consist of a static simulation experiment of a human model standing in a T-pose surrounded by eight cameras. In addition, some artificial occlusions were positioned to jeopardize the outcome of some of the cameras to test the triangulation robustness. The results are summarized in \tableref{static}. The first part of the table shows the results without occlusions, while the second one shows the results of the occlusions testing. The first column shows two pictures of the scene from different perspectives to appreciate the occlusions' effect. On the right, the table lists the MPJPE per joint with its standard deviation in three cases: i) not considering the weight (i.e., a standard Least Squares solution), ii) considering the confidence score provided by the 2D skeleton tracker, and iii) considering all the weights explained in detail in the \textit{Methods} section. The results show the excellent performance of the WLS method both in the case with and without occlusions. In the first example, when there are no occlusions, it is observed how the mean errors decrease from approximately 71~\si{mm} to approximately 10~\si{mm}. This is the most favorable case for performing triangulation, with the subject still and no occlusions. Our method allows a minimum mean error of less than half a centimeter for the right shoulder and with standard deviations not exceeding one centimeter in most cases, confirming our method's excellent performance and reliability. This fact becomes even more evident when a non-weighted approach and occlusions exist. In this case, the average error exceeds 400~\si{mm}, with the maximum error being more than 1~\si{m}. However, using the proposed approach, we get the average error to decrease to about 17~\si{mm}. That is, 20 times less than using an unweighted approach. One of the aspects to highlight from the results presented in this experiment is the significant influence of the score weight. This is also due to how the weights have been set, described in the \textit{Methods} section.

% \usepackage{multirow}
\begin{table}[t]
\begin{tabular}{ll|ccc|}
\cline{3-5}
                                         &                  & \multicolumn{3}{c|}{\textbf{MPJPE {[}\si{mm}{]}}}                                                                    \\ \cline{2-5} 
\multicolumn{1}{l|}{}                    & \textbf{Joints}  & \multicolumn{1}{c|}{\textbf{No Weights}} & \multicolumn{1}{c|}{\textbf{Score Weight}} & \textbf{All Weights}       \\ \hline 
\multicolumn{1}{|l|}{\multirow{13}{*}{}} & Neck             & \multicolumn{1}{c|}{$13.44 \pm 0.4$}             & \multicolumn{1}{c|}{$12.96 \pm 0.37$}                & $14.57 \pm 0.43$                     \\ \cline{2-5} 
\multicolumn{1}{|c|}{\multirow{2}{*}{\includegraphics[width=4.5cm]{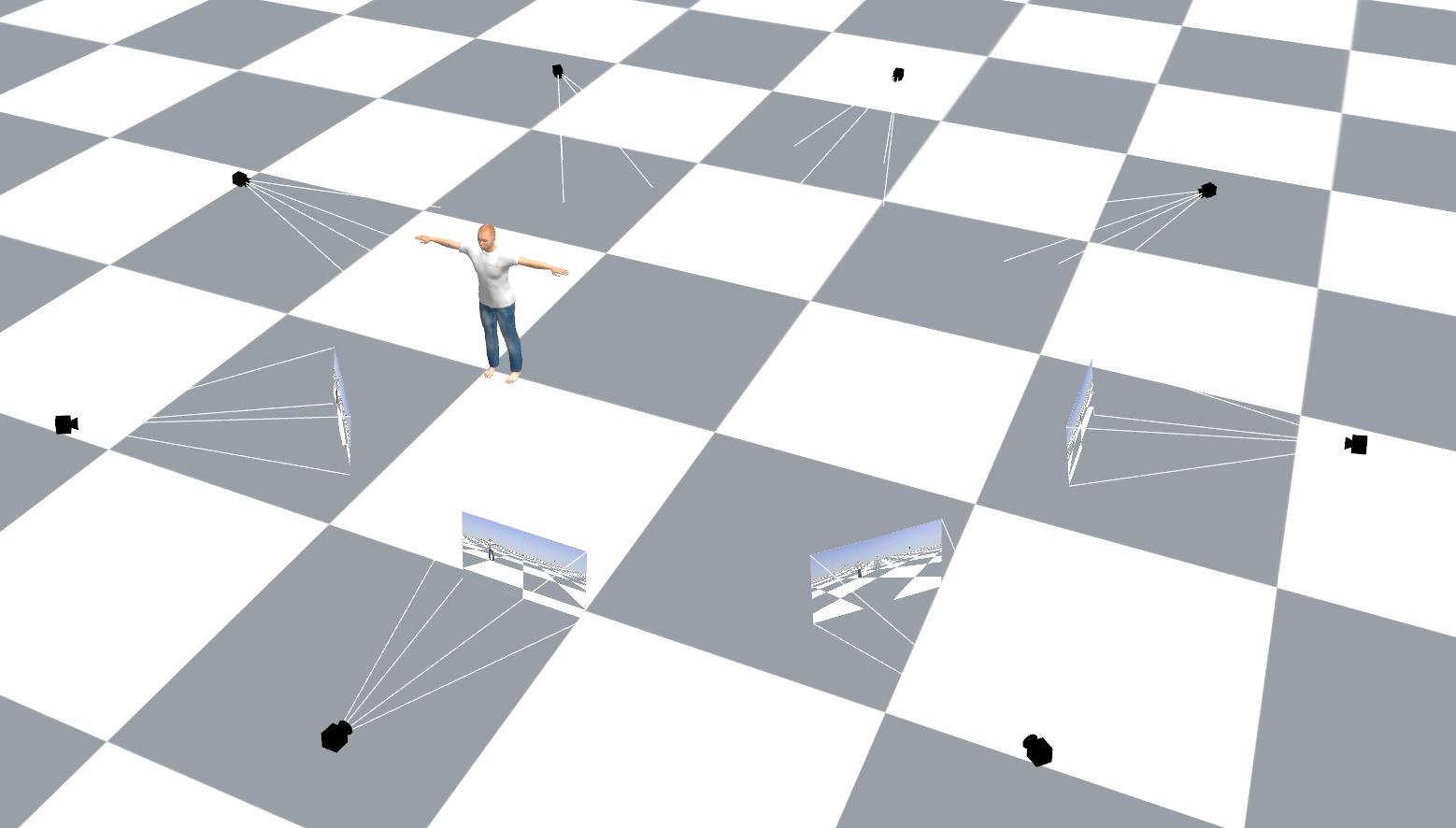}}}                   & Right Shoulder   & \multicolumn{1}{c|}{$3.77 \pm 0.60$}              & \multicolumn{1}{c|}{$3.70 \pm 0.64$}                & $4.51 \pm 0.71$                     \\ \cline{2-5} 
\multicolumn{1}{|l|}{}                   & Right Elbow      & \multicolumn{1}{c|}{$13.28 \pm 28.68$}            & \multicolumn{1}{c|}{$10.86 \pm 0.71$}                & $12.29 \pm 0.97$                     \\ \cline{2-5} 
\multicolumn{1}{|l|}{}                   & Right Wrist      & \multicolumn{1}{c|}{$145.72 \pm 282.33$}           & \multicolumn{1}{c|}{$14.47 \pm 0.62$}               & $13.27 \pm 0.76$                     \\ \cline{2-5} 
\multicolumn{1}{|l|}{}                   & Left Shoulder    & \multicolumn{1}{c|}{$11.95 \pm 0.50$}            & \multicolumn{1}{c|}{$10.28 \pm 0.40$}                & $11.80 \pm 0.52$                     \\ \cline{2-5} 
\multicolumn{1}{|l|}{}                   & Left Elbow       & \multicolumn{1}{c|}{$359.07 \pm 0.68$}            & \multicolumn{1}{c|}{$7.19 \pm 1.16$}                & $6.96 \pm 1.41$                     \\ \cline{2-5} 
\multicolumn{1}{|l|}{}                   & Left Wrist       & \multicolumn{1}{c|}{$376.84 \pm 7.82$}            & \multicolumn{1}{c|}{$14.66 \pm 0.76$}                 & $13.69 \pm 0.85$                       \\ \cline{2-5} 
\multicolumn{1}{|l|}{}                   & Hips             & \multicolumn{1}{c|}{$9.29 \pm 0.33$}            & \multicolumn{1}{c|}{$10.01 \pm 0.27$}                 & $9.12 \pm 0.40$                      \\ \cline{2-5} 
\multicolumn{1}{|l|}{\multirow{2}{*}{\includegraphics[width=4.5cm]{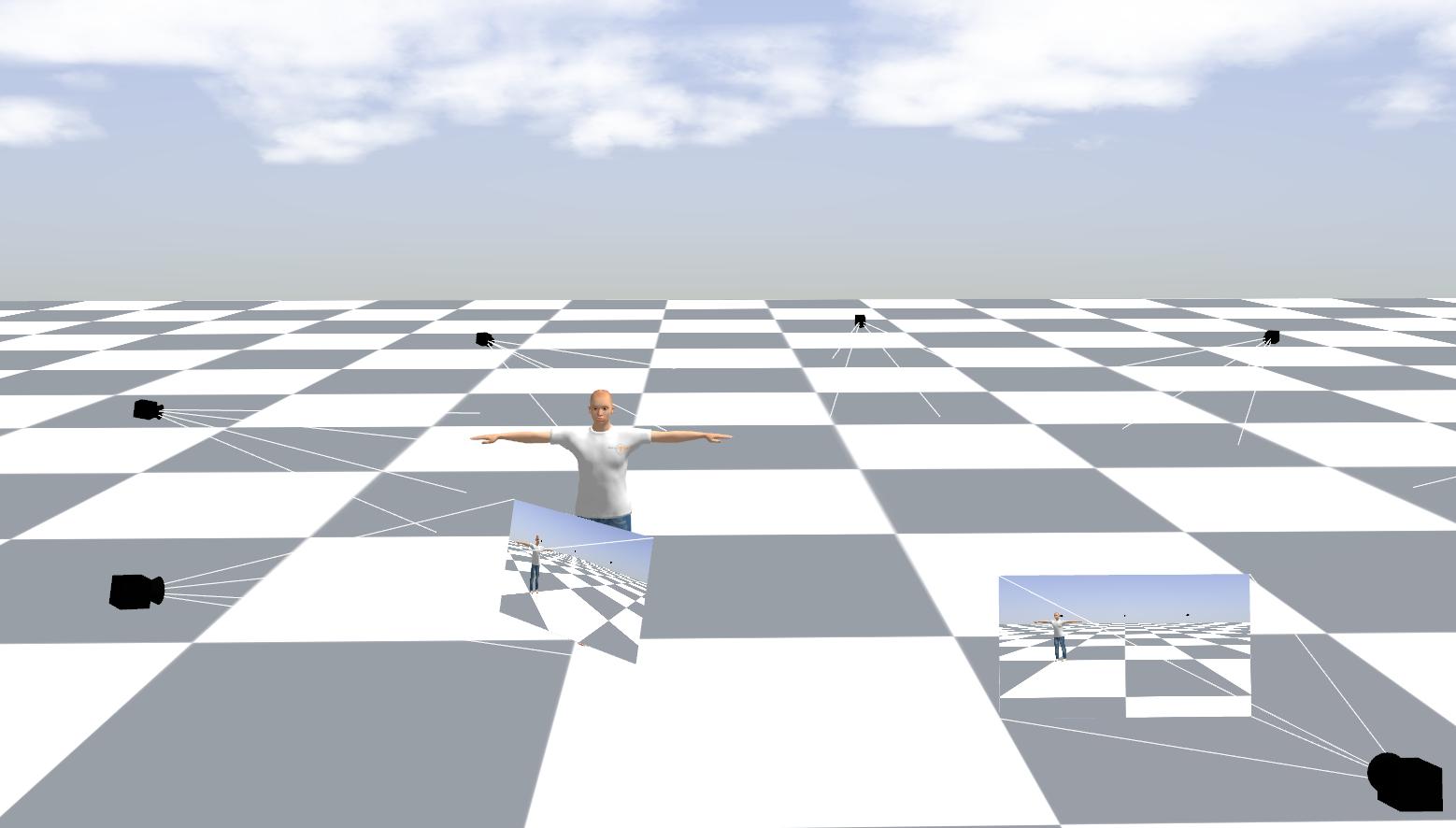}}}                   & Right Hip        & \multicolumn{1}{c|}{$12.14 \pm 0.58$}            & \multicolumn{1}{c|}{$14.33 \pm 0.60$}                 & $13.44 \pm 0.56$                      \\ \cline{2-5} 
\multicolumn{1}{|l|}{}                   & Right Knee       & \multicolumn{1}{c|}{$7.51 \pm 0.54$}            & \multicolumn{1}{c|}{$8.09 \pm 0.47$}                & $8.03 \pm 0.52$                     \\ \cline{2-5} 
\multicolumn{1}{|c|}{}                   & Right Ankle      & \multicolumn{1}{c|}{$13.01 \pm 0.41$}            & \multicolumn{1}{c|}{$12.98 \pm 0.31$}                 & $11.85 \pm 0.39$                      \\ \cline{2-5} 
\multicolumn{1}{|l|}{}                   & Left Hip         & \multicolumn{1}{c|}{$11.23 \pm 0.96$}            & \multicolumn{1}{c|}{$12.88 \pm 0.81$}                 & $11.98 \pm 1.06$                    \\ \cline{2-5} 
\multicolumn{1}{|l|}{}                   & Left Knee        & \multicolumn{1}{c|}{$7.39 \pm 0.69$}            & \multicolumn{1}{c|}{$6.55 \pm 0.51$}                 & $7.25 \pm 0.60$                      \\ \cline{2-5} 
\multicolumn{1}{|l|}{}                   & Left Ankle       & \multicolumn{1}{c|}{$8.91 \pm 0.66$}            & \multicolumn{1}{c|}{$9.15 \pm 0.63$}                & $7.92 \pm 0.70$                      \\ \cline{2-5} 
\multicolumn{1}{|l|}{}                   & \textbf{Average} & \multicolumn{1}{c|}{\textbf{70.97}}            & \multicolumn{1}{c|}{\textbf{10.58}}                & \multicolumn{1}{c|}{\textbf{10.48}} \\ \hline
\multicolumn{1}{|l|}{\multirow{13}{*}{}} & Neck             & \multicolumn{1}{c|}{$10.85 \pm 0.38$}             & \multicolumn{1}{c|}{$10.82 \pm 0.49$}                & $10.77 \pm 0.65$                     \\ \cline{2-5} 
\multicolumn{1}{|c|}{\multirow{2}{*}{\includegraphics[width=4.5cm]{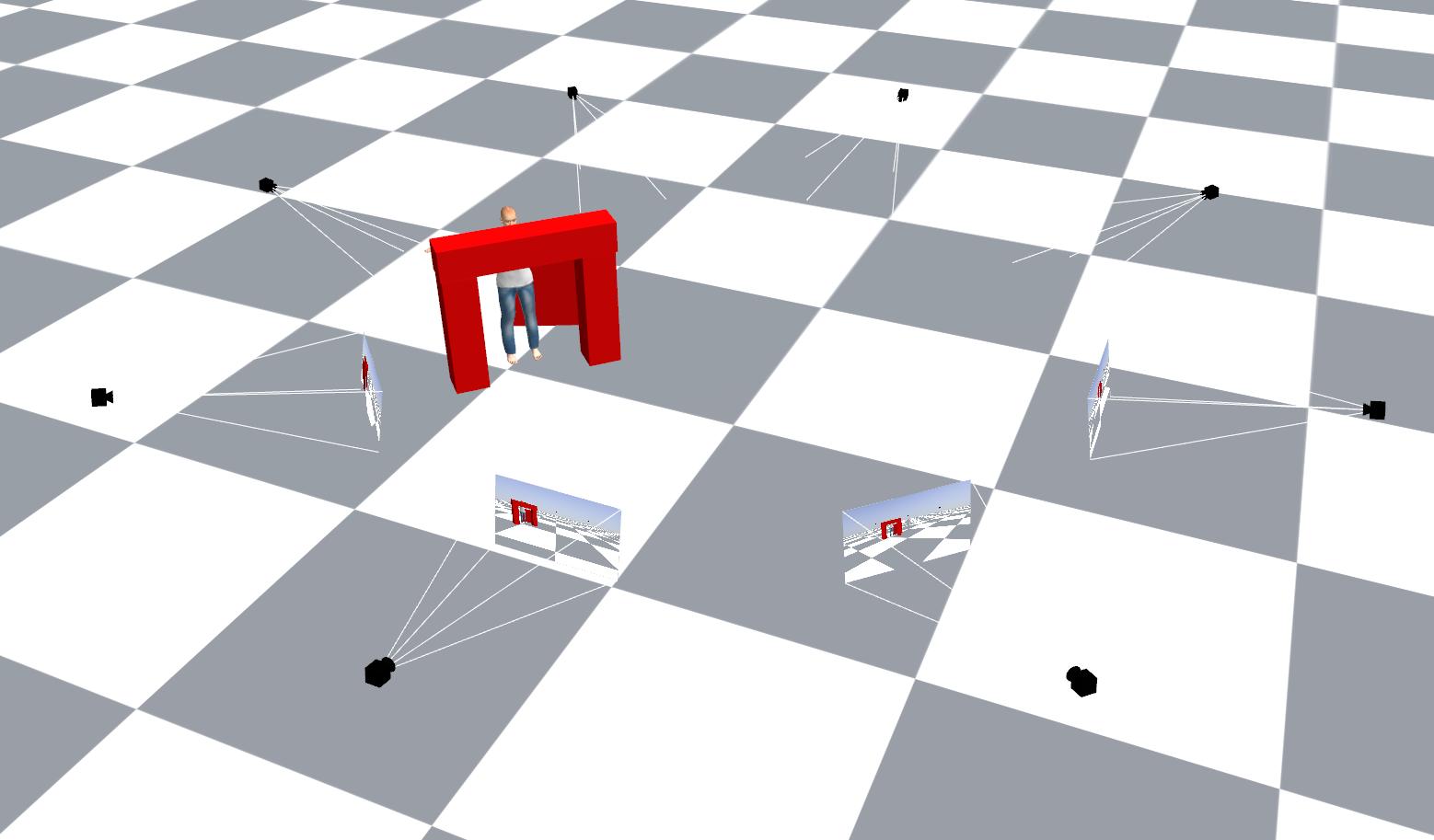}}}                   & Right Shoulder   & \multicolumn{1}{c|}{$9.20 \pm 0.68$}              & \multicolumn{1}{c|}{$22.15 \pm 0.86$}                & $12.23 \pm 0.78$                     \\ \cline{2-5} 
\multicolumn{1}{|l|}{}                   & Right Elbow      & \multicolumn{1}{c|}{$715.68 \pm 505.49$}            & \multicolumn{1}{c|}{$75.72 \pm 6.24$}                & $29.95 \pm 5.68$                     \\ \cline{2-5} 
\multicolumn{1}{|l|}{}                   & Right Wrist      & \multicolumn{1}{c|}{$1066.02 \pm 322.85$}           & \multicolumn{1}{c|}{$117.68 \pm 18.03$}               & $59.80 \pm 7.87$                     \\ \cline{2-5} 
\multicolumn{1}{|l|}{}                   & Left Shoulder    & \multicolumn{1}{c|}{$145.51 \pm 58.40$}            & \multicolumn{1}{c|}{$20.53 \pm 1.19$}                & $12.28 \pm 1.58$                     \\ \cline{2-5} 
\multicolumn{1}{|l|}{}                   & Left Elbow       & \multicolumn{1}{c|}{$220.33 \pm 23.73$}            & \multicolumn{1}{c|}{$51.87 \pm 3.64$}                & $39.43 \pm 3.38$                     \\ \cline{2-5} 
\multicolumn{1}{|l|}{}                   & Left Wrist       & \multicolumn{1}{c|}{$340.58 \pm 76.22$}            & \multicolumn{1}{c|}{$5.98 \pm 11.80$}                 & $6.08 \pm 2.44$                       \\ \cline{2-5} 
\multicolumn{1}{|l|}{}                   & Hips             & \multicolumn{1}{c|}{$410.09 \pm 1.94$}            & \multicolumn{1}{c|}{$8.20 \pm 0.69$}                 & $5.79 \pm 1.76$                      \\ \cline{2-5} 
\multicolumn{1}{|l|}{\multirow{2}{*}{\includegraphics[width=4.5cm]{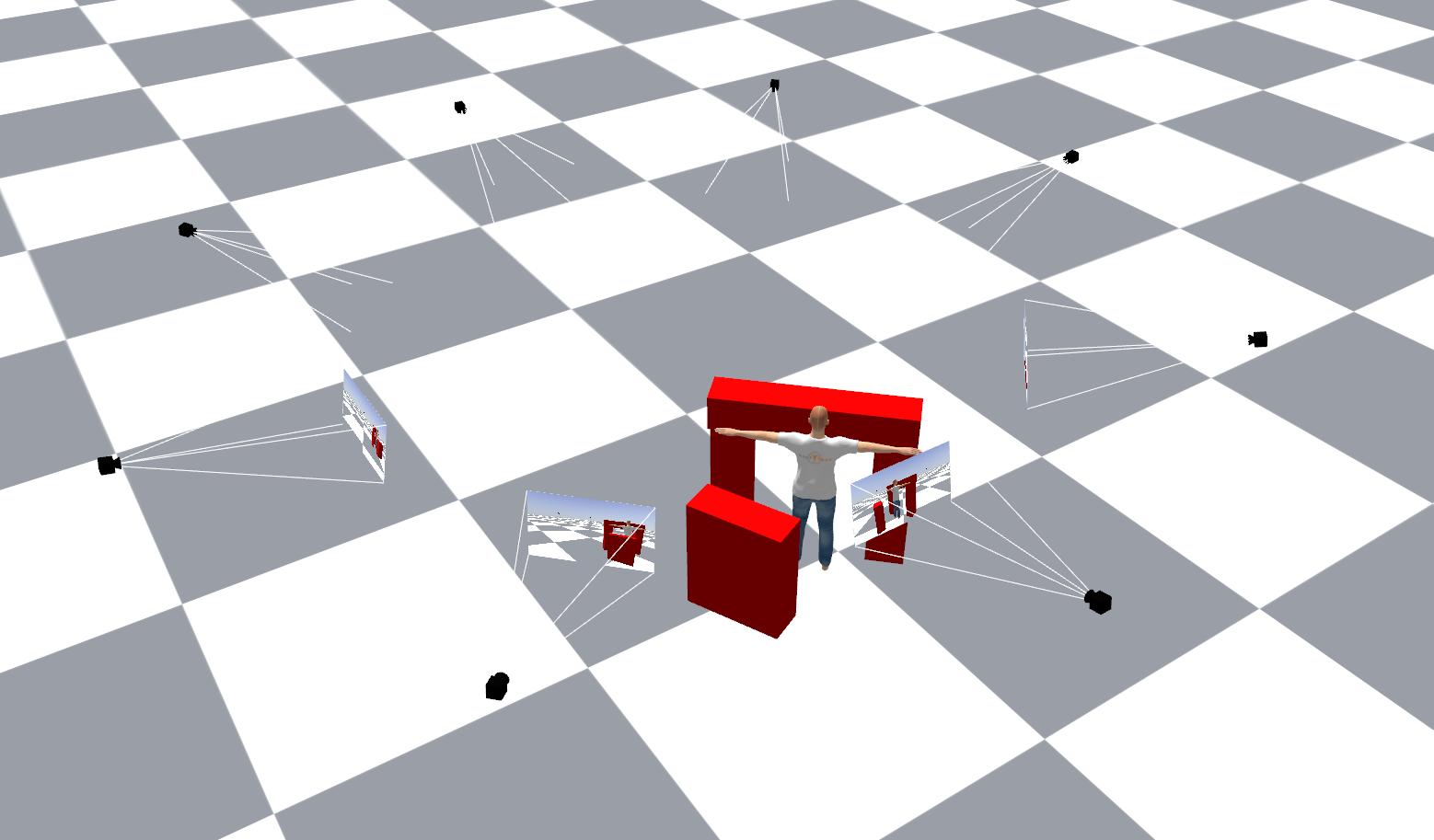}}}                   & Right Hip        & \multicolumn{1}{c|}{$491.48 \pm 13.41$}            & \multicolumn{1}{c|}{$9.01 \pm 2.28$}                 & $9.42 \pm 1.90$                      \\ \cline{2-5} 
\multicolumn{1}{|l|}{}                   & Right Knee       & \multicolumn{1}{c|}{$548.00 \pm 20.48$}            & \multicolumn{1}{c|}{$11.47 \pm 0.68$}                & $12.92 \pm 0.96$                     \\ \cline{2-5} 
\multicolumn{1}{|c|}{}                   & Right Ankle      & \multicolumn{1}{c|}{$602.63 \pm 30.25$}            & \multicolumn{1}{c|}{$8.96 \pm 1.79$}                 & $5.75 \pm 1.29$                      \\ \cline{2-5} 
\multicolumn{1}{|l|}{}                   & Left Hip         & \multicolumn{1}{c|}{$498.04 \pm 30.20$}            & \multicolumn{1}{c|}{$7.68 \pm 1.20$}                 & $16.99 \pm 2.22$                    \\ \cline{2-5} 
\multicolumn{1}{|l|}{}                   & Left Knee        & \multicolumn{1}{c|}{$548.94 \pm 40.28$}            & \multicolumn{1}{c|}{$9.25 \pm 0.78$}                 & $8.83 \pm 1.19$                      \\ \cline{2-5} 
\multicolumn{1}{|l|}{}                   & Left Ankle       & \multicolumn{1}{c|}{$600.34 \pm 52.75$}            & \multicolumn{1}{c|}{$10.95 \pm 0.82$}                & $9.04 \pm 0.36$                      \\ \cline{2-5} 
\multicolumn{1}{|l|}{}                   & \textbf{Average} & \multicolumn{1}{c|}{\textbf{443.40}}            & \multicolumn{1}{c|}{\textbf{26.44}}                & \multicolumn{1}{c|}{\textbf{17.09}} \\ \hline
\end{tabular}
\captionsetup{labelfont=bf,font=sf}
\caption{\textbf{\textbf{Static assessment of the weights effect in the 3D reconstruction}.} The table shows two scenarios of a subject standing in an 8-cameras surrounded workspace in two situations: one with no obstacles and one with some randomly generated obstacles. The cameras are positioned on the perimeter of  4.5 \si{m} radius circumference. For each case, the average MPJPE error with its variability is calculated considering three cases: No Weights, Score Weight, and All Weights. Finally, the values are averaged among all key points to summarize the outcome of the results.}
\label{table:static}
\end{table}

\subsection*{\textsf{3D human pose tracking: Real-time performance evaluation}}

\begin{figure}[t!]
    \centering
    \includegraphics[width=0.85\columnwidth]{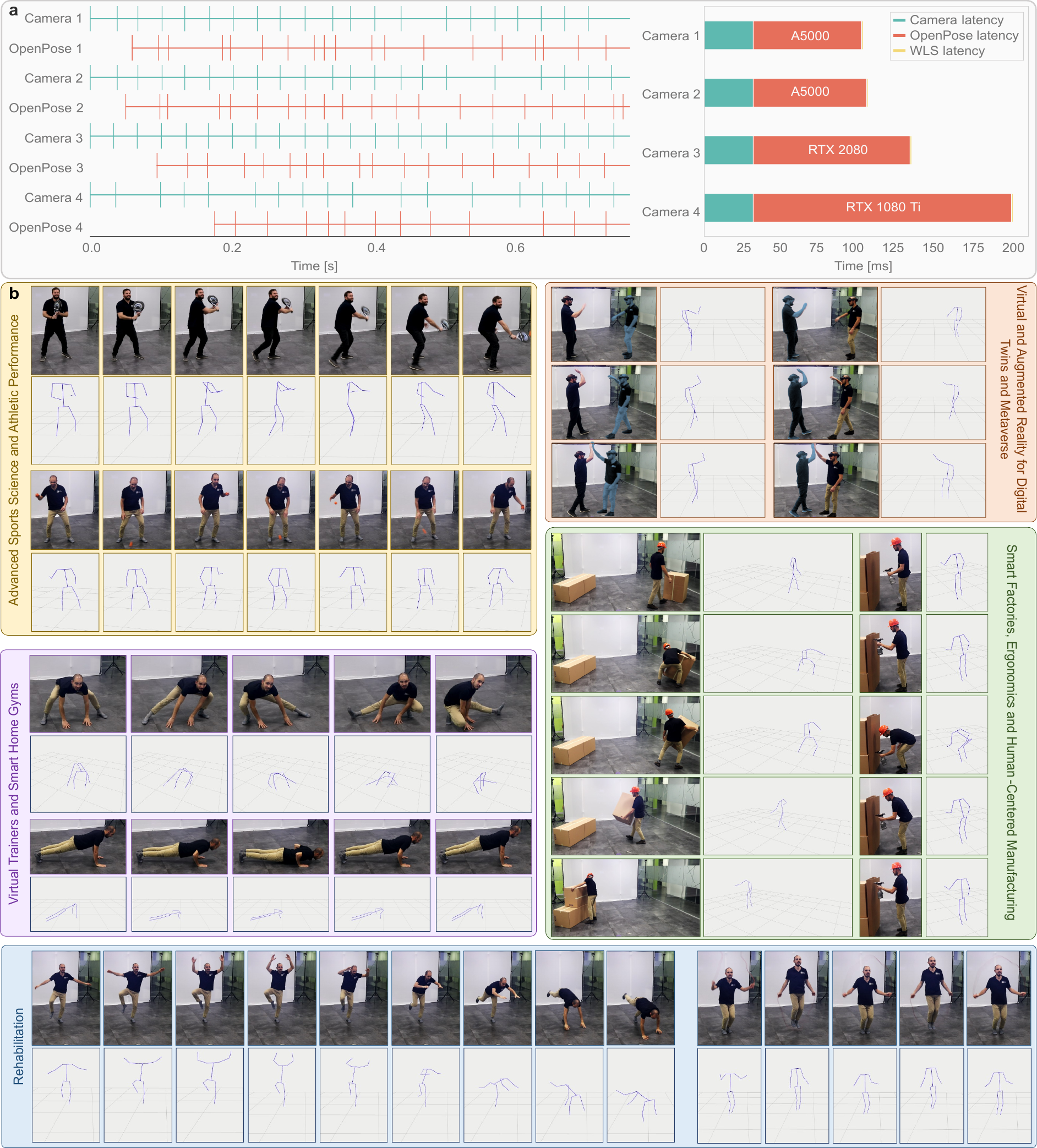}
    \captionsetup{labelfont=bf,font=sf}
    \caption{\textbf{An illustration of the framework's potential deployment in real applications with real-time requirements}. A latency analysis is depicted at the top of the figure. The graph on the left is a spike plot showing in green the timestamp at which a frame streamed by a camera is read from the OpenPose node, and in red, the timestamp when the estimated frame is outputted. On the right, the latency is averaged for every camera over each, also showing the effects of using different GPU models. The GPU models used are written on top of the red bars in the right-hand plot. A further discussion on the computer network setup is carried out in the \textit{Methods} section. The excellent performance of 3D reconstruction, both in terms of accuracy and real-time, makes it eligible for a wide variety of tasks. Examples of these tasks using the proposed methodology are shown at the bottom of the figure.}
    \label{fig:applications}
\end{figure}

A latency analysis has been conducted to quantitatively demonstrate the real-time capabilities of the proposed framework. Latency analysis is the process of measuring the delay or latency in a system. It involves identifying the bottlenecks and sources of delay and determining the root causes of those delays.
The analysis is conducted in the same experimental setup described in the previous section, with four cameras framing a person walking around. Note that the subject's movement details and the environment type (i.e., real-world or simulation) are irrelevant to the results, as only the computation time of the different subprocesses matters. 
The results are depicted in \figref{applications}a. 
The capture rate of the cameras is set to 30~\si{Hz} (i.e., the sensors will capture thirty images every second: 30~\si{FPS}). Hence, a capture latency of $33.3 \si{ms}$ is introduced at this step. 
Moving to the pipeline's next step, we analyze the latency introduced by the 2D human pose tracker (i.e., OpenPose). The spike plot in the left graph of \figref{applications}a summarizes the results. Our framework allows an abstraction of the hardware in the sense that it allows using different cameras and GPUs. However, the computational or time performance of the 2D human pose trackers available is highly dependent on the computational power of the GPU. This experiment uses three different GPUs (see section Methods for more details). As shown in the right plot of \figref{applications}a, the average latency can vary from 25 to 125 \si{ms} depending on the GPU. The last step of the method pipeline is the triangulation that, as can be appreciated in the same plot, shows a $<1 \si{ms}$ latency. This analysis shows clearly that the bottleneck is due to 2D tracking methods. However, thanks to the integration of the method with Open-VICO in ROS, the 3D triangulation method always runs at 100Hz, and the output is updated every time new data is received from one of the cameras asynchronously. Evidently, this can generate some errors that will be larger the more dynamic the movements are with respect to the total latency of the system. A more detailed discussion on this aspect is presented in the \textit{Discussion} section.

In Fig.~\ref{fig:applications}~\textbf{a}, various individuals are shown in various situations, performing different movements with their corresponding virtual avatars next to them. The individuals do not wear any physical markers or sensors, but their movements are tracked and recorded by the proposed markerless framework. Overall, this figure exhibits our proposal's versatility and potential applications with real-time requirements, which can be used to track and analyze a wide range of human movements, from athletic performance to medical rehabilitation.

\subsection*{\textsf{Comparative analysis with a commercial system}}
As described in the introduction, marker-based commercial systems are not the ideal solution for testing a tracking system due to the problems already illustrated. That is why the qualitative comparison of the accuracy of the tracking method was performed in a simulation environment where the ground truth is wholly known. However, to claim that our framework can be used in real applications, we must conduct a comparative analysis with a commercial system. This comparison is included in this section.

\begin{figure}[]
    \centering
    \includegraphics[width=0.75\columnwidth]{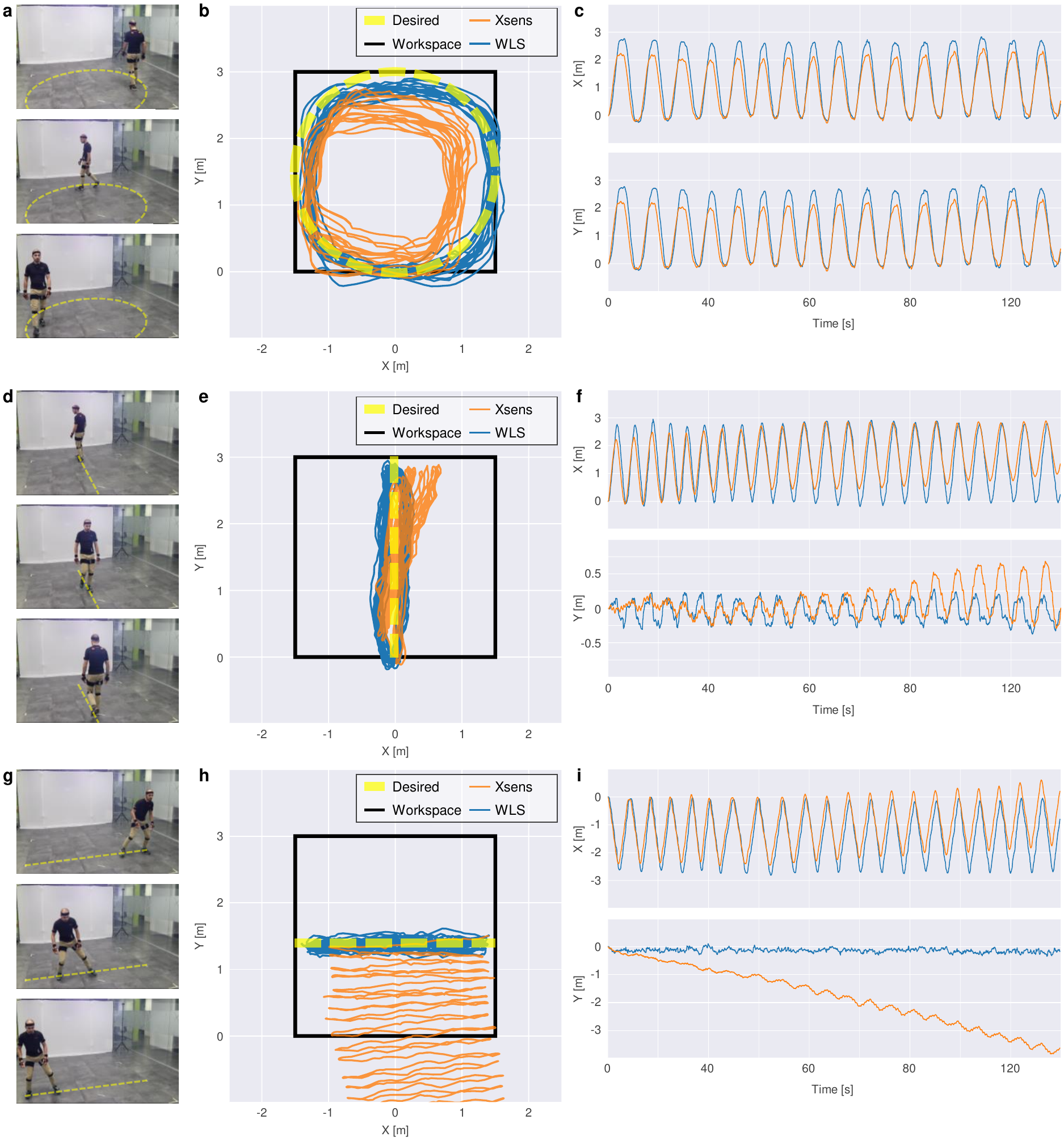}
    \captionsetup{labelfont=bf,font=sf}
    \caption{\textbf{Performance comparison of the proposed triangulation framework with a wearable commercial IMU-based system.} The figure can be macroscopically be divided in two experiments. On the top the results of some occlusion driven examples are represented. On the bottom a drift analysis is shown comparing the recording of the horizontal (left) and vertical (right) position of the pelvis landmark over time of the two systems. Over all the images from top to bottom the color orange is used to indicate Xsens data while the blue one indicates the proposed weighted least square methodology. (\textbf{a}) 2D views of a half-body frontal occlusion. (\textbf{c}) 2D views of a half-body frontal occlusion with fake arm color consistent with the shirt. (\textbf{b}-\textbf{d}) 3D overlapping of the two systems. (\textbf{e}) Circular path. (\textbf{h}) Back and forth path. (\textbf{k}) Lateral sliding path. (\textbf{f}-\textbf{i}-\textbf{l}) Horizontal (XY) position of the pelvis of a person walking along predefined path. (\textbf{g}-\textbf{j}-\textbf{m}) X and Y components variation over time.} 
    \label{fig:paths}
\end{figure}

Here we present a qualitative comparison of the system evaluated against a commercial marker-based system to highlight the strengths and weaknesses of the two. Optical-based motion capture systems are often compared with inertial-based systems. While the former may be susceptible to occlusion when markers are blocked from view, they generally do not suffer from drifting problems commonly associated with inertial-based systems. Occlusion occurs when a person or object being tracked is partially or fully obstructed from view by another object in the scene, usually resulting in a loss of tracking information for the occluded body part or object, making it difficult to estimate its position, orientation, or movement accurately. Optical marker-based systems are highly affected by this issue. Drift occurs when the measurements of acceleration and rotation accumulate small errors over time, resulting in a gradual deviation from the actual position and orientation of the body segments. Inertial-based systems are particularly vulnerable to this problem, which can be especially problematic for long captures or complex movements involving motion and direction changes. On the other hand, existing AI-based markerless techniques are based on systems that classify pixels as body parts. One of the problems with learning-based systems, in general, is that they do not consider uncertainties and, when they fail, can result in meaningless outputs even if the classification score is high. For the specific case of human pose estimation, they can result in utterly wrong body parts.

Two experiments have been carried out where these problems take place. In the first one, occlusions and tricky color-matching body parts occur, allowing us to evaluate our system's robustness against this phenomenon in real-world applications. In the second one, different walking patterns illustrate that our system does not present a drifting problem. The elected device for the comparison is Xsens (Enschede, Netherlands), one of the leading manufacturers of IMU technology. The Xsens system comprises 17 IMU sensors attached along the body that communicate wirelessly to an access point. The reason for using this system for comparison is because it was available in the laboratory and because it can be used in both experiments, the drift problem being visible in the second one. The difficulty of calibrating both systems and the impossibility of having an actual ground truth prevents us from comparing the two systems quantitatively. That is, we could obtain the MPJPE of our system and the commercial system, but we could not know whether this error is due to possible calibration errors between both systems, errors in the tracking of our proposal, or those of the commercial system. Therefore, we could only evaluate these experiments by a qualitative analysis.

The top of \figref{paths} reflects the results of the first experiment. \figref{paths}a-d show the outcome of two explanatory cases demonstrating the consistency of our proposal. 
In particular, the proposed system exhibits superior performance to Xsens in accurately matching body pose, based on a qualitative evaluation of \figref{paths}a and \figref{paths}b.

In the second portion of \figref{paths}, we analyze the drifting issue in a standard working environment. The test is performed in a $6\times7 \si{m^2}$ room not shielded from magnetic disturbances to acquire insight into the method's absolute position determination during its use in a standard environment. Three walking paths at an average walking speed of between $3$ and $4 \si{km/h}$ with the same start and end points are set, as shown in \figref{paths}e,~h,~k. The first consists of a circle with a $3 \si{m}$ diameter, the second is a $3 \si{m}$ back-and-forth walking, and the last is a $3 \si{m}$ lateral sliding. 
The paths in the $XY$ plane are plotted in a ``scatter diagram'' (\figref{paths}f,~i,~l) to show how this issue affects our proposal in comparison with the commercial system. This plot represents the top view of the location of the pelvis landmark, which is in the middle of the body and suffers the least from small pose changes than joints in arms and legs.
The $X$ and $Y$ trajectories of this joint are depicted in \figref{paths}g,~j,~m respectively for each walking path.
While drifting is appreciable in all cases of the marker-based system, the results show the robustness of the proposed framework to this issue. The graphs in \figref{paths}l-m (i.e., lateral sliding path) stands out with the highest drift along the Y axis for the marker-based system, whereas our proposal is not affected by this issue. In addition, one can notice that this is just a short recording; thus, longer sessions using an IMU-based system would result in much larger errors caused by drifting.

% SECTION 4: Discussion
\section*{\textsf{Discussion}}
\label{Sec::discussion}
As introduced at the beginning of this manuscript, the development of a markerless real-time 3D human pose tracking system poses several challenges. One of the main ones is the need to integrate robust and accurate algorithms for human motion tracking in real applications, which requires the development of methods that can effectively handle a wide range of human poses and movements while being resistant to error sources. A second challenge is the need to develop efficient and scalable implementations of these algorithms. This is especially important for real-time tracking, where the ability to process large amounts of data quickly is critical. A third challenge is the need to integrate tracking algorithms with other system components, such as sensors, robots, or other elements that may vary depending on the application, which requires the development of efficient interfaces and protocols to enable the flow of data and commands between the various components. The results presented in this manuscript support the conclusion that the proposed framework presents solutions to all these challenges and represents a complete solution for developing a markerless real-time 3D human pose tracking system that can be easily applied in numerous applications. In addition, it does not require complex calibration systems, demonstrating its readiness for real applications outside the laboratory. 

Although using multi-camera systems with 2D AI-based tracking systems is not novel per se, this approach with a WLS method that allows the integration of different error sources to minimize position error is. So is its integration in a comprehensive framework that allows its use in real-world applications, with a simple pipeline that integrates a simple and accurate method of calibration of the multi-camera system, along with the possibility of integrating different hardware components in a straightforward way increasing flexibility, or being able to integrate different 2D AI-based human tracking methods. In addition, thanks to Open-VICO, the possibility of integrating it in robotics applications with ROS and allowing the passage from simulation to the real world and vice versa in a simple way increases the possibility of being used in automation and robotics applications, with as many cameras as needed to cover all the workspace required.

The results reported in the \textit{Accuracy analysis} experiments are significant and promising, especially considering that open-source pose estimation algorithms (e.g., OpenPose) are not designed for biomechanical applications, and the datasets they have been trained on are, theoretically, inaccurate in detecting joint centers. Despite this, the results report very high accuracy rates, with low errors even in highly challenging situations with occlusions. In particular, the experiment with occlusions presented in the section \textit{comparative analysis with a commercial system} has also been attempted with a marker-based commercial optical system. However, this system could not work as the occlusions impeded performing an adequate reconstruction of the human skeleton, which is why it was not introduced in the experiment. The fact that our system presents accuracies of around one centimeter, even in the presence of occlusions, is a living demonstration of the effectiveness of the proposed methodology.

The latency analysis reported the results introduced into the system by the various framework components. It has been found that the bottleneck is in the 2D pose estimators, which are computationally heavy. One of the limitations of our framework is its dependency on 2D human pose estimators that are, at the same time, highly hardware-dependant. 
Nevertheless, this bottleneck latency can be reduced using more powerful hardware or lighter 2D estimators. Also, using cameras with higher frequencies could reduce the camera capture latency. Thanks to the software and hardware abstraction of our framework, high-frequency cameras, computers with higher computational power, or lighter 2D tracking systems could be easily integrated, allowing not only to work in real-time but also at high frequencies. This fact demonstrates the scalability of the proposed framework, especially for highly dynamic applications.

The fact that the system allows human movements to be integrated into simulation environments is a major step forward in creating human digital twins.
By using human digital twins to test motion capture algorithms, we can assess how well the algorithms perform in capturing the movements of a human being. They can also use the digital twin to create a range of scenarios, such as different lighting conditions or environmental factors, to test the algorithm's robustness and accuracy.
In addition, using human digital twins to test motion capture algorithms can help identify areas for improvement and refine them, which can have a wide range of applications, including sports analysis, virtual reality, and gaming. Overall, using human digital twins in testing motion capture algorithms can improve the accuracy and reliability of these algorithms and create more realistic and immersive virtual environments. 

Our novel approach for real-time tracking of 3D human motion has significant potential to impact various fields and advance human society. In healthcare, our system could be used for remote monitoring and analysis of patient movement. In sports, it could provide coaches and athletes with detailed, real-time analysis of movement and technique. In entertainment, it could enable the creation of immersive, interactive experiences. Our framework is robust, efficient, and accurate and has the potential to revolutionize the way we interact with technology and understand human movement. In the subsequent works, we intend to extend the system to a real-time multi-target framework without influencing accuracy and real-time performance. For this purpose, we will be inspired by existing works that try to solve the feature-matching problem between different cameras and the transition from 2D to 3D spaces. This is a contribution of high interest for the scalability of the proposal of this manuscript that deserves to be studied independently and is outside the scope of this work.

% SECTION 5: Methods
\section*{\textsf{Methods}}
\label{Sec::methods}
This section describes the complete application pipeline, from installing the cameras to acquiring the necessary data, either in simulation or in the real environment. In particular, the calibration method of the cameras to obtain the extrinsic parameters is described. Next, the human body representation and the method of 3D tracking of the human skeleton are described. Finally, we describe the setup used in the real and simulation environments that allowed us to obtain the results presented above.

\subsection*{\textsf{Extrinsic calibration of the multi-camera system}}
\label{sec::calibration}

First, it is necessary to place the cameras and calibrate them. Given the characteristics of our framework, this step can be performed in the real world or simulation. This section describes the former, as the latter does not require calibration, and details on how it can be done can be found in the Open-VICO article. One aspect to be considered at this point is that it is necessary to use at least three cameras sharing a common workspace in which a person whose body, or most of it, can be seen by all the cameras. The number of cameras needed depends on the application and the required working space. Our framework allows the placement of as many cameras as necessary, being limited by the computational capacity of the hardware used, as discussed in the discussion section.

Once the cameras have been placed, it is necessary to carry out an extrinsic calibration of them to get the relative poses between them and to establish a global reference system $\mathcal{W}$ called world.
Extrinsic calibration of a set of cameras is a process extensively covered in the literature, and many solutions have been proposed. The problem consists of assessing the poses of a set of static cameras, and the solution consists in obtaining references that can be matched between the different cameras. The most commonly used procedures in the literature typically use 2D calibration patterns~\cite{zhang2000flexible}. 

The main issue is that the visibility of the pattern may be limited, and it must be moved to different locations to be observed by all cameras. Then, all the information must be combined afterward. In this work, we used the approach developed in~\cite{sarmadi2019simultaneous}, which is handy in calibrating multi-camera setups with a shared field of view as ours. The article proposed a method that, given a synchronized video sequence showing an object comprised of a set of squared planar markers moving freely in front of the cameras, automatically estimates the 3D structure of the planar markers (calibration prop), the 3D poses of the cameras, and the relative pose between the object and the cameras. 
This method has multiple hallmarks that make it the best pick. First, compared to a classic method, the calibration prop can be viewed by more cameras simultaneously, optimizing all global extrinsic parameters simultaneously. While in the case of a chessboard or asymmetric circle grid, calibrations are done pairwise, and the global extrinsic parameters are inferred from those. The second reason is that the calibration process can be done without knowing the calibration prop shape and configuration. 

\begin{figure}
    \centering
    \captionsetup{labelfont=bf,font=sf}
    \includegraphics[width=0.85\columnwidth]{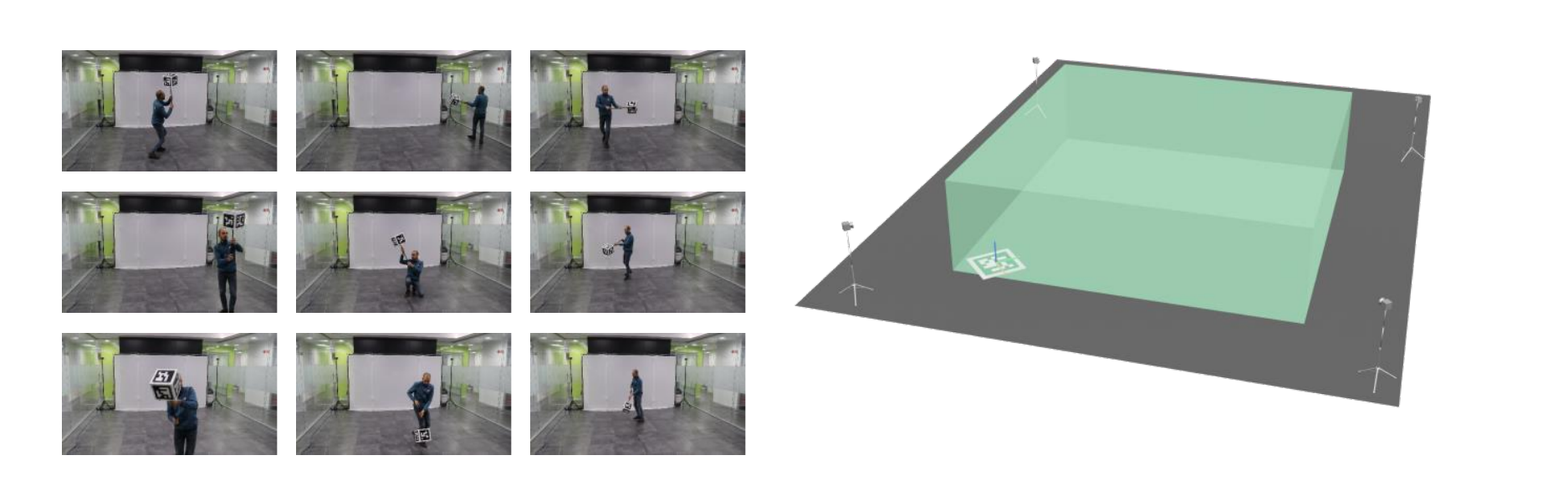}
    \caption{\textbf{Calibration routine.} The left picture shows photos of the wanding procedure using a custom ArUco-covered cube as a calibration prop moved around the workspace. On the right, one can see a rendered representation of the calibrated environment with an ArUco tile to fixing the world reference frame.}
    \label{fig:calibration}
\end{figure}

The calibration process followed is shown in Fig.~\ref{fig:calibration}. A 3D cubic calibration prop realized with ArUco markers is waved in the capture volume. The time of this process vary depending on the number of cameras and the volume of the workspace. In the particular case shown in this work the process took around two minutes. The images are collected and synchronized using their ROS timestamps, allowing us to apply the method proposed in~\cite{sarmadi2019simultaneous}. Once the extrinsic calibration problem is solved, the poses of the cameras can be transformed into the world reference frame by using an additional ArUco marker tile on the ground. Note that it is not the purpose of this work to show the accuracy of the calibration method since it has been extensively covered in~\cite{sarmadi2019simultaneous} and~\cite{fortini2022openvico}. 
However, to confirm that the calibration was satisfactory, one can check the position w.r.t. $\mathcal{W}$ of an additional ArUco marker or the point cloud overlapping if using RGB-D or stereo cameras. The latter is the one used in this work. 

\subsection*{\textsf{Simulation environment in Gazebo}}
As previously described, the framework was tested in simulation and real-world environments. 
The simulation was implemented in a Gazebo environment, utilizing specific user-friendly tools provided by Open-VICO. Open-VICO provides a variety of tools that enable the creation of different camera setups using easily-writable \textit{yaml} files. Additionally, it offers the ability to create and animate realistic human models that stream their skeleton information in a ROS network. This feature is particularly beneficial for prototyping vision-based human tracking algorithms and makes Open-VICO an essential toolbox for this purpose. A camera model using a simple camera plugin was used to conduct the experiments. Different cameras configuration (i.e., number of cameras and positions) were used depending on the experiment. The experiments were carried out using a single PC, recording the data singularly for every camera, not to be limited by GPU. Then, the data were processed offline with a workstation (CPU: Intel\textsuperscript{\textregistered}
 Xeon\textsuperscript{\textregistered} W-2245 Processor, GPU: 2xNVIDIA\textsuperscript{\textregistered} RTX A5000, RAM: 4x16~\si{GB} Hynix\textsuperscript{\textregistered} DDR4 3200).

\subsection*{\textsf{Real-world experimental setup}}
The real-world experimental setup consists of four RealSense D435 RGB-D cameras positioned on each corner of a $6\times7 \si{m^2}$ room to maximize the workspace's coverage. Note that the use of depth information is not required for the execution of the proposed methodology. We decided to use RGB-D cameras for their availability and ease of integration in a ROS environment. The devices are connected to a computer network consisting of 3 computers to maximize GPU and CPU performance. The computers present the following computational features
\begin{itemize}
	\item Computer 1 $\rightarrow$ 2xIntel\textsuperscript{\textregistered} Realsense D435, CPU: Intel\textsuperscript{\textregistered} Xeon\textsuperscript{\textregistered} W-2245 Processor, GPU: 2xNVIDIA\textsuperscript{\textregistered} RTX A5000, RAM: 4x16~\si{GB} DDR4 3200
	\item Computer 2 $\rightarrow$ 1xIntel\textsuperscript{\textregistered} Realsense D435, CPU: Intel\textsuperscript{\textregistered} Core\textsuperscript{\texttrademark} i7-10750H, GPU: NVIDIA\textsuperscript{\textregistered} GeForce RTX 2080, RAM: 2x16~\si{GB} DDR4 2667
	\item Computer 3 $\rightarrow$ 1xIntel\textsuperscript{\textregistered} Realsense D435, CPU: Intel\textsuperscript{\textregistered} Core\textsuperscript{\texttrademark} i7-7700, GPU: NVIDIA\textsuperscript{\textregistered} GeForce GTX 1080, RAM: 2x16~\si{GB} DDR4 2400
\end{itemize}
Due to the standardized communication protocol between nodes, the distributed system operates under a ROS network, simplifying communication between cameras and pose-tracking algorithms, particularly when changing camera models or human pose detection libraries. The cameras are calibrated using the automated routine described before. 

\begin{figure}[t]
    \centering
    \includegraphics[width=1\columnwidth]{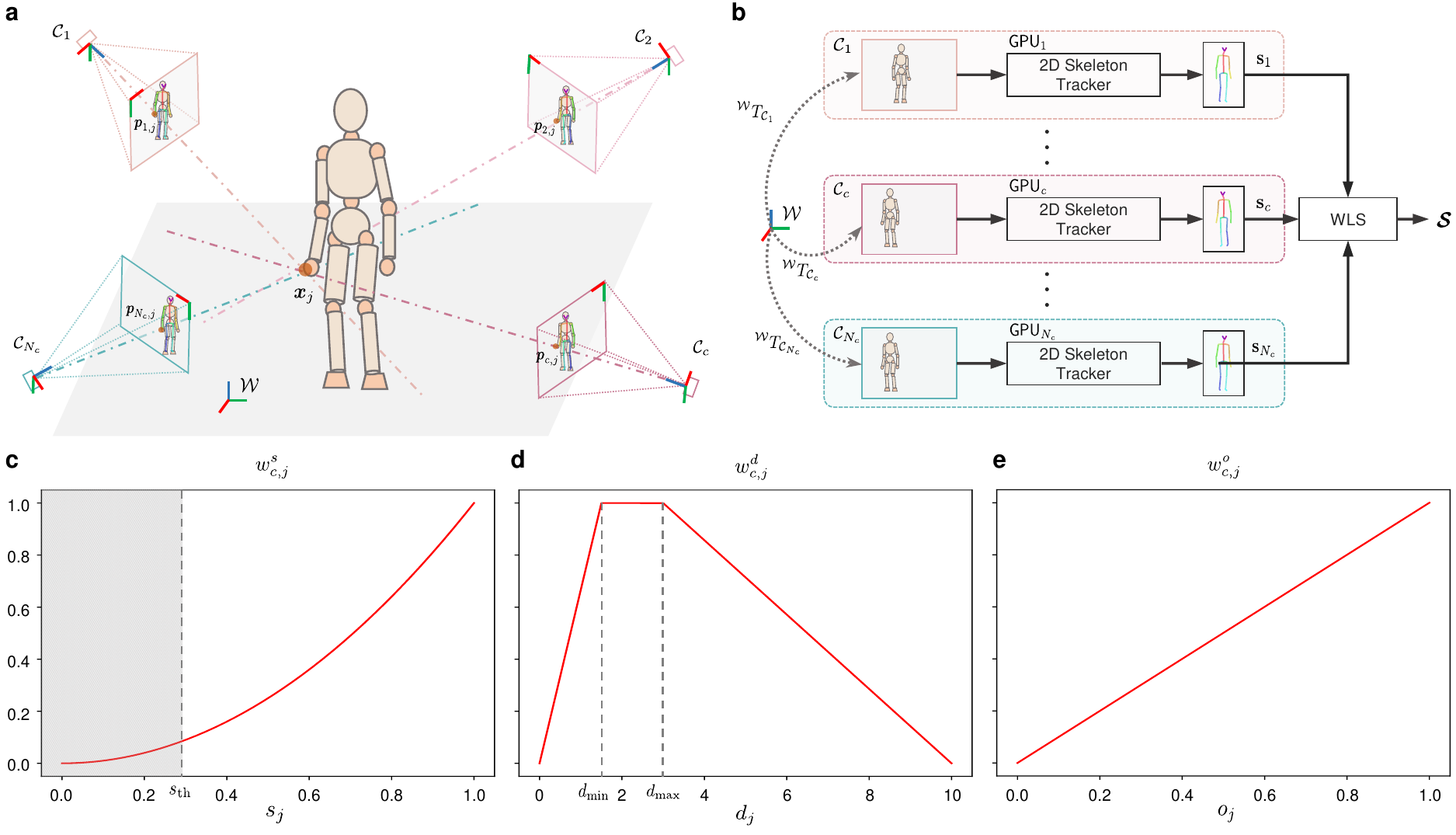}
    \captionsetup{labelfont=bf,font=sf}
    \caption{\textbf{Schematic illustration of the proposed triangulation method}. \textbf{(a)} Model example of a multi-camera system where each camera obtains a human skeleton in the camera plane. The specific case of the projection of the right wrist joint $\boldsymbol{p}_{c,j}\in\mathbb{R}^2$ from the 2D planes of each camera to the 3D world is represented. The WLS allows the computation of the respective joint $\boldsymbol{x}_{j}\in\mathbb{R}^3$  w.r.t. the $\mathcal{W}$ frame. \textbf{(b)} Schematic of the method pipeline. Each camera feeds the WLS method with a set of 2D joints obtained employing a 2D human-pose tracker. Knowing the extrinsic parameter matrix $^WT_{\mathcal{C}_c}$ that allows computing the homogeneous transformations of each camera in the world frame $\mathcal{W}$, a skeleton $\boldsymbol{\mathcal{S}}$  can be constructed in $\mathcal{W}$. \textbf{(c)-(e)} Functions representing the choice of the three weights chosen (score $w_{s,j}$, distance $w_{d,j}$, and orthogonality $w_{o,j}$) for the WLS method.}
    \label{fig:method}
\end{figure}

\subsection*{\textsf{Multi-camera triangulation for 3D human skeleton tracking}}
The proposed image-based 3D reconstruction aims to create a real-time three-dimensional kinematic representation of the human body from a set of images. The cameras capture the same scene from different viewpoints, then the parallax of the images is obtained to calculate the 3D coordinate~\cite{julesz1960binocular, bi2020multicamera}. 
In traditional 3D reconstruction using multi-camera systems, the camera's perspective projection matrix converts 2D image coordinates into 3D world coordinates. Methods such as least squares or normalized least squares are used to solve the equation systems, with each camera contributing equally to the final 3D reconstruction accuracy. However, several factors might affect the performance of the 3D reconstruction, introducing errors in the system. In a multi-camera setting, several errors source might be induced by multiple factors, such as camera distance, resolution, orientation, and occlusions. Hence, the quality can be inconsistent across the cameras, potentially reducing the overall accuracy of conventional methods. 

This article proposes a weighted 3D reconstruction approach that employs a Weighted Least Squares (WLS) process, allowing us to ponder each camera's contribution to each body part for an accurate reconstruction. Following this procedure, a more accurate reconstruction can be achieved by decreasing the weight assigned to cameras with more significant errors and increasing the weight assigned to smaller ones.

The fundamental concept of our multi-camera approach is illustrated in \figref{method}. Let us define a human skeleton model $\boldsymbol{\mathcal{S}} = \{\boldsymbol{x}_1, \ldots, \boldsymbol{x}_j, \ldots , \boldsymbol{x}_{N_j}\}$ composed of $N_j$ landmarks (i.e., joints). Each $\boldsymbol{x}_j\in\mathbb{R}^3$ represents the 3D position of each joint $j$ w.r.t. the world reference frame $\mathcal{W}$. Let us also define a set of $N_c$ cameras, where each camera $c$ is placed in a 3D environment in a way that a certain volume of the environment is captured by all the cameras. Hence, every point of interest placed inside this volume can be seen by all the cameras. \figref{method}a represents this system, where from each camera, we get a 2D representation of the human body using an AI-based skeleton tracking system. Open-VICO allows us to use different methods at this point; in this work, we use OpenPose, for being one of the most widespread in the literature. The output is a 2D skeleton model $\boldsymbol{s}_c$ for each camera $c$ w.r.t. the camera frame $\mathcal{C}_c$ defined by a set of $N_j$ joints. Each joint $\boldsymbol{p}_{c,j}\in\mathbb{R}^2$ represents the position of the joint in the $\mathcal{C}_c$ frame that can then be projected to $\mathcal{W}$ using the projection matrix $\boldsymbol{M}_c\in\mathbb{R}^{4\times4}$ according to a pinhole camera model.
\begin{equation}
\label{eq:pinhole}
z_c \left[\begin{array}{c}
    \boldsymbol{p}_{c,j}\\
    1
    \end{array}\right] = 
    \boldsymbol{M}_c \left[\begin{array}{c}
    \boldsymbol{x}_{j}\\
    1
    \end{array}\right]
\end{equation}
where $z_{c,j}$ is the unknown depth information of the joint $j$ w.r.t. the camera $c$. The projection matrix $\boldsymbol{M}_c$ is computed with the intrinsic and extrinsic matrices according to
\begin{equation}
\label{eq:projection_matrix}
\boldsymbol{M}_c= \boldsymbol{\mathcal{I}}_{c} \, ^\mathcal{W}\boldsymbol{T} _{\mathcal{C}_c}^{-1} 
\end{equation}
where $\boldsymbol{\mathcal{I}}_{c}\in\mathbb{R}^{3\times4}$ is the intrinsic matrix of the $c$-th camera, which can be computed by calibrating each camera independently or using the one given by the camera provider. The latter is the approach followed in this work. $^\mathcal{W}\boldsymbol{T} _{\mathcal{C}_c}\in\mathbb{R}^{4\times4}$ is the extrinsic matrix $c$-th camera obtained from the calibration routine described above. These two matrices are defined as
\begin{align}
    & \boldsymbol{\mathcal{I}}_{c} = \left[\begin{array}{cccc}
    f_{c,x} & 0 & k_{c,x} & 0 \\
    0 & f_{c,y} & k_{c,y} & 0 \\
    0 & 0 & 1 & 0
    \end{array}\right] 
	& ^\mathcal{W}\boldsymbol{T} _{\mathcal{C}_c} =\left[\begin{array}{cc}
	\begin{array}{cc}
	^\mathcal{W}\boldsymbol{R}_{\mathcal{C}_c}
	\end{array} & ^\mathcal{W}\boldsymbol{t}_{\mathcal{C}_c} \\
	\boldsymbol{0}_{1\times3} & 1
\end{array}\right] 
\end{align}
where $f_{c,x}$ and $f_{c,y}$ are the pixel focal lengths, and are identical for square pixels. The $k_{c,x}$ and $k_{c,y}$ values are the offsets of the principal point, from the top-left corner of the image frame. $^\mathcal{W}\boldsymbol{R}_{\mathcal{C}_c}\in\mathbb{R}^{3\times3}$ and $^\mathcal{W}\boldsymbol{t}_{\mathcal{C}_c}\in\mathbb{R}^{3\times3}$ are the rotation and translations of $\mathcal{C}_c$ w.r.t. $\mathcal{W}$. $\boldsymbol{0}_{1\times3}$ is an horizontal vector with 3 elements that are all zeros.

The \equref{pinhole} cannot be solved as the depth information of the detected joints $z_{c,j}$ is unknown. However, employing a multi-camera approach allows us to eliminate this component and express a system of equations in the form
\begin{equation}
\label{eq:problem_single}
    \boldsymbol{A}_j \boldsymbol{x}_j = \boldsymbol{B}_j
\end{equation}
where $\boldsymbol{A}_j \in \mathbb{R}^{2 N_c \times 3}$ and $\boldsymbol{B}_j\in \mathbb{R}^{2N_c}$ are computed as 
\begin{align}
\label{eq:A_B_matrices}
    & \boldsymbol{A}_j=\left[\begin{array}{c}
    \boldsymbol{P}_{1,j} \left[\begin{array}{cc} \boldsymbol{m}_{1,j}^{{(3)}}-\boldsymbol{m}_{1,j}^{{(1)}} & \boldsymbol{0}_{1\times4} \\ \boldsymbol{0}_{1\times4} & \boldsymbol{m}_{1,j}^{{(3)}}-\boldsymbol{m}_{1,j}^{{(2)}} \end{array}\right] \\ \\
    \vdots \\ \\
    \boldsymbol{P}_{c,j} \left[\begin{array}{cc} \boldsymbol{m}_{c,j}^{{(3)}}-\boldsymbol{m}_{c,j}^{{(1)}} & \boldsymbol{0}_{1\times4} \\ \boldsymbol{0}_{1\times4} & \boldsymbol{m}_{c,j}^{{(3)}}-\boldsymbol{m}_{c,j}^{{(2)}} \end{array}\right] \\ \\
    \vdots \\ \\
    \boldsymbol{P}_{N_c,j} \left[\begin{array}{cc} \boldsymbol{m}_{N_c,j}^{{(3)}}-\boldsymbol{m}_{N_c,j}^{{(1)}} & \boldsymbol{0}_{1\times4} \\ \boldsymbol{0}_{1\times4} & \boldsymbol{m}_{N_c,j}^{{(3)}}-\boldsymbol{m}_{N_c,j}^{{(2)}} \end{array}\right] \\
    \end{array}\right]
    &  \boldsymbol{B}_j=\left[\begin{array}{c}
    \left[m_{1,j}^{(1,4)}, \, m_{1,j}^{(2,4)}\right]- m_{1,j}^{(3,4)} \boldsymbol{P}_{1,j}  \\ \\
    \vdots \\ \\
    \left[m_{c,j}^{(1,4)}, \, m_{c,j}^{(2,4)}\right]- m_{1,j}^{(3,4)} \boldsymbol{P}_{c,j}  \\ \\
    \vdots \\ \\
    \left[m_{N_c,j}^{(1,4)}, \, m_{N_c,j}^{(2,4)}\right]- m_{1,j}^{(3,4)} \boldsymbol{P}_{N_c,j}
    \end{array}
    \right]
\end{align}
where $\boldsymbol{m}_{c,j}^{(r)}\in\mathbb{R}^4$ denotes $r$-th row of the $c$-th camera projection matrix $\boldsymbol{M}_c$; and $m_{c,j}^{(r,l)}\in\mathbb{R}$ denotes the element in the $r$-th row, $l$-th column of the $c$-th camera projection matrix $\boldsymbol{M}_c$. Note that here the joint $\boldsymbol{p}_{c,j}$ is written in the matrix form
\begin{equation}
    \boldsymbol{P}_{c,j}=\left[\begin{array}{cc}
    p_{c,j}^{(1)}  & 0 \\ 
    0 & p_{c,j}^{(2)}
    \end{array}
    \right]
\end{equation}
where $p_{c,j}^{(r)}\in\mathbb{R}$ denotes the $r$-th element of $\boldsymbol{p}_{c,j}$ in the joint $j$ of camera $c$.

A solution to the system of equations in~\ref{eq:problem_single} for each joint $j$ solves our problem, allowing us to obtain a 3D reconstruction of a human skeleton from multiple 2D skeletons. As discussed above, the solution to this problem is addressed using a WLS-based approach, represented in Fig.~\ref{fig:method}~\textbf{b}, where the WLS is fed by the 2D skeletons provided by each camera. Hence, the problem described in equation~\ref{eq:problem_single} is formulated as a WLS problem in the form 
\begin{equation}
\label{eq:triangulation}
    \boldsymbol{x}_j=[\boldsymbol{A}_j^\top\boldsymbol{W}_j\boldsymbol{A}_j]^{-1}\boldsymbol{A}_j^\top\boldsymbol{W}_j\boldsymbol{B}_j
\end{equation}
where the weighting matrix $\boldsymbol{W}_{j}\in\mathbb{R}^{2N_c \times 2N_c}$ per joint is defined as a square diagonal matrix of the weighting matrices per camera. 
\begin{equation}
\label{eq:weight_matrix}
    \boldsymbol{W}_j = \textrm{diag}\left(\begin{bmatrix} 
    w_{1,j} & 0  \\
    0 & w_{1,j} 
    \end{bmatrix}, \cdots, \begin{bmatrix} 
    w_{c,j} &  0 \\
    0 & w_{c,j} 
    \end{bmatrix}, \cdots, \begin{bmatrix} 
    w_{N_c,j} & 0  \\
    0 & w_{N_c,j} 
    \end{bmatrix}\right)
\end{equation}  
where $w_{c,j}$ is the individual weight associated to the camera $c$ and joint $j$, computed according to
\begin{equation}
    \label{eq:individual_weight}
    w_{c,j} = w_{c,j}^s\left(\frac{w_{c,j}^o + w_{c,j}^d}{2}\right)
\end{equation}
In a WLS problem, the weights assigned determine the relative importance or influence on the final solution based on different factors. This approach aims to minimize the sum of the weighted squared residuals, where the residuals are the differences between the observed data points and the predicted values from the model. The weights are typically chosen to reflect the precision or reliability of the observations so that data points with higher precision or reliability are given more weight in the fit. The weights of Equation~\ref{eq:weight_matrix} are determined empirically, and their profile can be appreciated in Fig.~\ref{fig:method}~\textbf{c-e}. $w_{c,j}^s$ is the square of the confidence score $s_j$ OpenPose assigns to every joint according to the quality outcome of the estimation. Weighting the residuals by the square of the weights can help to reduce the impact of outliers or data points with large errors that might otherwise have an undue influence on the fit. However, when the confidence score of joints is low (e.g., below 0.4), we noticed that it is better to disregard these points in the triangulation process. This threshold is represented as $s_{th}$ in Fig.~\ref{fig:method}~\textbf{c}. $w_{c,j}^d$ weights the distance of the $j$-th joint wrt the $c$-th camera following a linear profile with its maximum in the interval $d_{min} < d_j <d_{max}$ as suggested by~\cite{zago2000tracking}. Finally, $w_{i,j}^o$ is called the orthogonality index and gives a measure of the orientation of the body w.r.t. $\mathcal{C}_c$, quantifying the visibility as the inverse of the scalar product of the optical axis of the camera with the Z-axis of the body part (described in the following section). 

\subsection*{\textsf{Human skeletal model representation}}
\label{sec::frames}
\begin{figure}
    \centering
    \captionsetup{labelfont=bf,font=sf}
    \includegraphics[width=0.85\columnwidth]{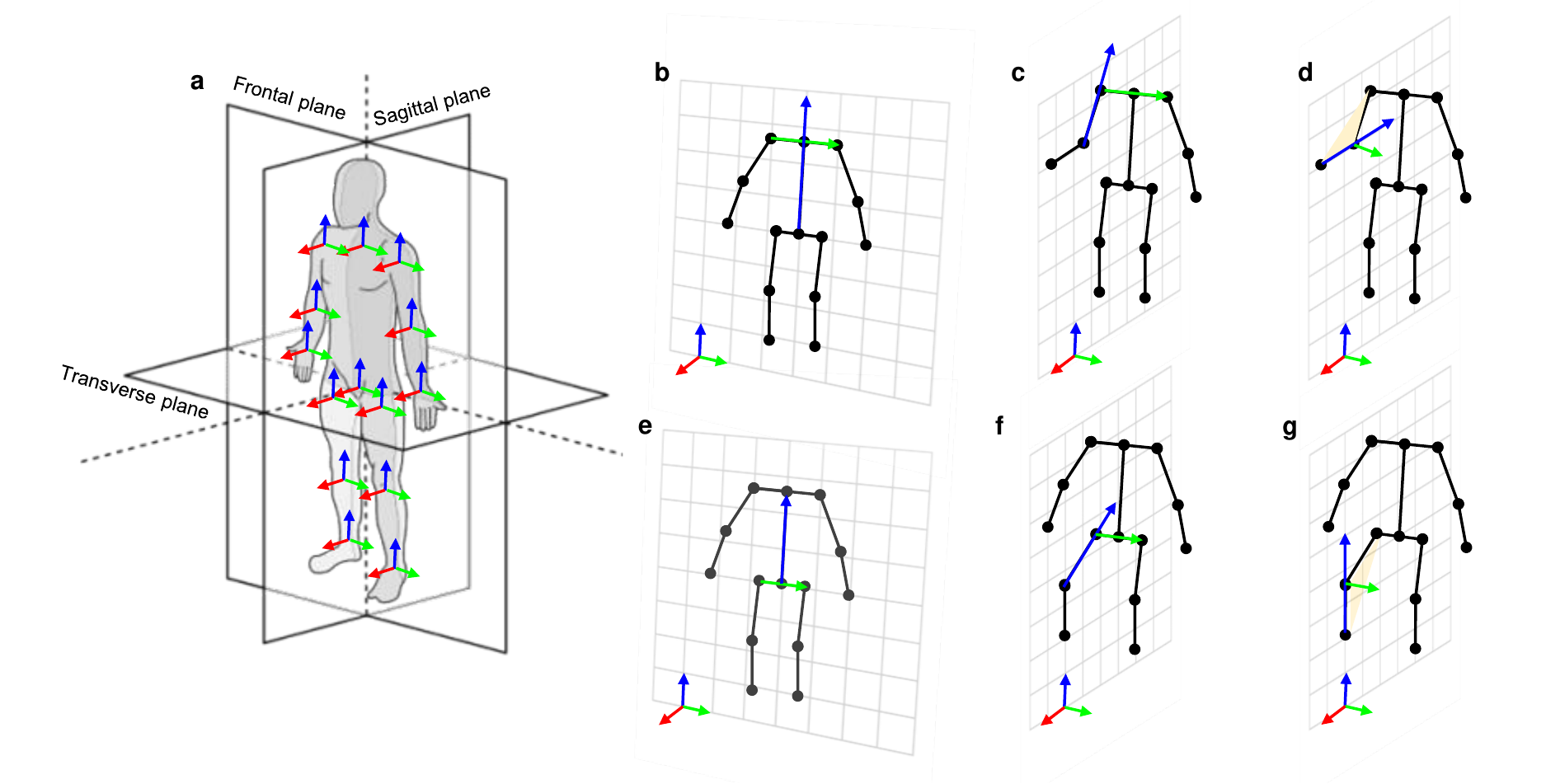}
    \caption{\textbf{References conventions used to compute the Links Coordinate Systems.} (\textbf{a}) Anatomical planes (frontal, sagittal, and transversal) . (\textbf{b}) Chest frame. (\textbf{c}) Upper arm frame. (\textbf{d}) Forearm frame. (\textbf{e}) Pelvis frame. (\textbf{f}) Upper leg frame. (\textbf{g}) Lower leg frame. For every link when an axis is not represented means that is calculated as the cross product of the others.}
    \label{fig:frames}
\end{figure}

The final skeleton reconstruction process was performed as follows. First, the 3D joint locations were extracted using the proposed triangulation method. Two successive joints are connected to represent the body segments. The Joint Coordinate System (JCS), according to the  International Society of Biomechanics (ISB)~\cite{wu1995isb, wu2002isb, wu2005isb} standard, was then computed providing 3D coordinates for each joint based on their anatomic characteristics. With this information, the angle measure for each segment is calculated. The movement angle could then be input to control the interactive application. The segment and the axes are then used to compute the plane angle, which is the one between the segment and the normal of the plane where it should be occurring. The references used to compute the JCS recommended by ISB are presented in Fig~\ref{fig:frames}. Before starting to analyze the joints, the following assumptions should be made. For the upper body, the Z-axis is used to perform a change of coordinates, which is represented by a vector from the pelvis center to the neck, defined as the midpoint of the shoulders' key points (see Fig.~\ref{fig:frames}~\textbf{b}). In the ISB convention, the X-axis is oriented anteriorly and away from the body. To determine the X-axis, the cross-product between the Y-axis and the scapular girdle (shoulder-to-shoulder vector) is computed. The same principle is applied to the upper arm, with the Z-direction, elbow-shoulder distance vector, and Y-axis serving as the normal vector to the plane defined by the forearm and upper arm vectors (see Fig.~\ref{fig:frames}~\textbf{c}). The X-axis is again computed as the cross-product of the Z and Y axes. Moving down the upper limb, the forearm link is encountered, with the Z-direction defined by the wrist-elbow vector distance and the Y-direction determined by the orthogonal vector of the plane defined by the forearm and the upper arm (see Fig.~\ref{fig:frames}~\textbf{d}). The Pelvis frame is defined by two vectors, the Y-direction obtained as the hips distance and the Z-direction as the neck-pelvis length(see Fig.~\ref{fig:frames}~\textbf{e}). Moving to the lower limb keypoints the Z-direction of the upper leg is obtained as the knee-hip distance while the lower leg is the ankle-knee length. The Z-direction of the Upper leg link and the lower leg one determine a plane whose normal vector is the Y-direction of both links. the X-direction of both these links is as a consequence the vectorial product of the other two directions (see Fig.~\ref{fig:frames}~\textbf{f} and Fig.~\ref{fig:frames}~\textbf{g}).

\printbibliography

\section*{\textsf{Acknowledgments}}
This work was supported in part by the European Union’s Horizon 2020 research and innovation program under Grant Agreement No. 871237 (SOPHIA) in part by the ERC-StG Ergo-Lean (Grant Agreement No.850932).

\section*{\textsf{Author contributions}}
The author contributions according to the CRediT Taxonomy are the followings:
Conceptualization: L.F., M.L., J.G.
Data curation: L.F., M.L.
Formal Analysis: L.F., J.G.
Funding acquisition: A.A.
Investigation: L.F., M.L., J.G.
Methodology: L.F., M.L.
Project administration: J.G., A.A.
Resources: A.A
Software: L.F., M.L.
Supervision: J.G., E.M., A.A
Validation: L.F., M.L., J.G.
Visualization: L.F., J.G.
Writing – original draft: L.F., J.G
Writing – review \& editing: L.F., M.L., J.G, E.M., A.A
\end{document}